\documentclass[11pt]{article}

\usepackage[final]{acl}

\usepackage{times}
\usepackage{latexsym}

\usepackage[T1]{fontenc}

\usepackage[utf8]{inputenc}

\usepackage{microtype}

\usepackage{inconsolata}

\usepackage{graphicx}

\usepackage{verbatim}
\usepackage{listings}
\usepackage{xcolor}
\usepackage{booktabs}
\usepackage{multirow}
\usepackage{array}
\usepackage{subcaption}
\usepackage{fontawesome}
\usepackage{lipsum}
\usepackage{amsmath}
\usepackage{longtable}
\usepackage{xurl}

\title{How Loud Rumbles Hit Newsstands: A Data Analysis of Coverage and Spatial Bias in German News about Landslides Around the World}

\author{\\
  \textbf{Brielen Madureira\textsuperscript{1,2}},
  \textbf{Andreas Niekler\textsuperscript{1,3}},
  \textbf{Marc Keuschnigg\textsuperscript{4,5}},
  \textbf{Mariana Madruga de Brito\textsuperscript{2}}
\\\\
  \textsuperscript{1}\small LeipzigLab -- Climate Discourse, Leipzig University, Germany \hfill
  \textsuperscript{2}\small Helmholtz Centre for Environmental Research, Germany \\
  \textsuperscript{3}\small Computational Humanities, Leipzig University, Germany \hfill
  \textsuperscript{4}\small Institute of Sociology, Leipzig University, Germany \\
  \textsuperscript{5}\small Institute for Analytical Sociology, Linköping University, Sweden
\\
  \small{
    \textbf{Correspondence:} \href{mailto:brielen.madureira@uni-leipzig.de}{brielen.madureira@uni-leipzig.de}
  }
}

\begin{document}
	
\maketitle

\begin{abstract}
	Landslides often hit newsstands due to their destructive and potentially fatal effects. News are a valuable source of information for creating or enriching disaster databases and for expediting media-based studies of the dynamics of media attention. To accomplish that, news datasets must be filtered, geolocated and validated. This paper focuses on how landslides around the world are reported in German newspapers. We analyse almost 55k news articles about 4.5k news events in a 25-year period, compare it with external measures of countries' susceptibility to landslides and provide insights, e.g.~the overreporting of Southern and Western Europe, to foster further studies on inequalities in media attention to international disasters.
\end{abstract}

\section{Introduction}
\textit{Sounds of cracking or breaking wood, groaning of the ground, a loud rumble and the sensation of a passing freight train}. Poetic as they may sound, these words are used by the U.S.~Geological Survey\footnote{\scriptsize\url{www.usgs.gov/programs/landslide-hazards/what-are-signs-landslide-development-what-do-i-do-if-a-landslide-occurs}} to describe the warning signs of a hazardous type of event: landslides. Also known as mass movements, they are ``the downslope movement of soil, rock, and organic materials under the effects of gravity'', e.g.~rockfalls and debris flow, triggered by water, seismic, volcanic or human activity \citep{highland2008landslide} and also influenced by climate change \citep{Gariano2022}. They can cause vast socioeconomic impact and fatalities, demanding mitigation and adaptation measures \citep{Kjekstad,Petley2012}.

\begin{figure}[ht]
	\centering
	{%
		\setlength{\fboxsep}{2pt}%
		\setlength{\fboxrule}{0.8pt}%
		\fbox{\includegraphics[trim={0 5.5cm 0 0},clip,width=0.85\columnwidth]{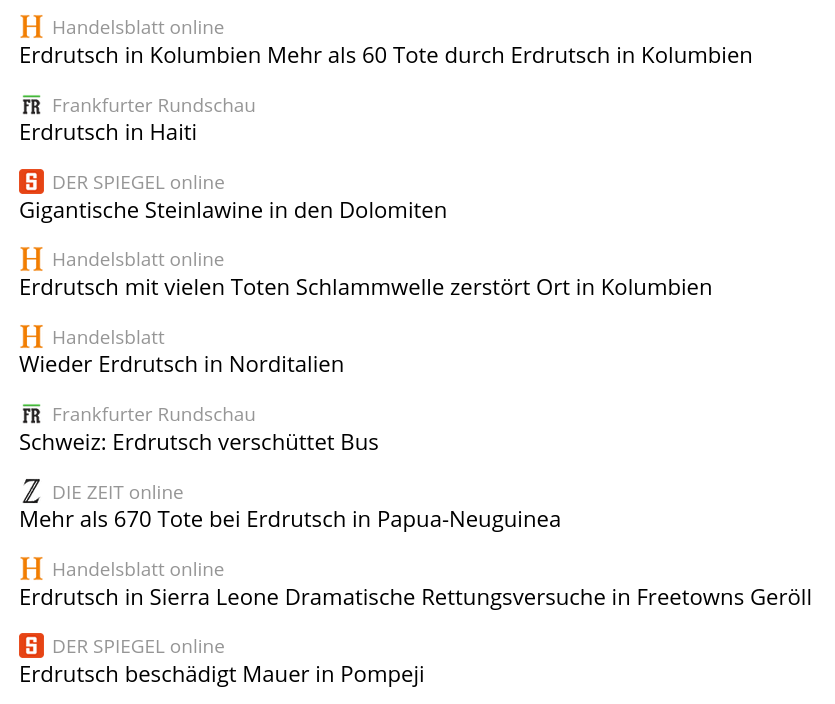}}
	}%
	\caption{Headlines about landslides around the world in German news from the \texttt{wiso-net} database.}
	\label{fig:news-search}
\end{figure}

Historical data on the occurrence and impact of landslides and other disasters worldwide are available in dedicated inventories such as EM-DAT \citep{Delforge2025} and the Global Landslide Catalog (GLC) \citep{Kirschbaum2009}. However, global databases exhibit spatial biases, limited temporal coverage, missing data and specific inclusion criteria that lead to incompleteness \cite{gall2009,Jones2023}. For instance, GLC contains only rainfall-triggered landslides and EM-DAT has strict inclusion thresholds (e.g.~at least 10 fatalities) and relies on limited data sources.

News have emerged as an alternative source of disaster information (see Section \ref{sec:lit}) but they stem from latent criteria \citep{Galtung1965,stgaard1965,stijn,Alipour2024}, such as the affected country's proximity or narrative appeal. Disproportions in media attention to landslides arise from the interplay of at least two factors: how often such disasters occur in reality and the newsworthiness of each event and country, given implicit media choices. 

Datasets of news about disasters thus serve two purposes: (i) enriching existing disaster databases within a text-as-data paradigm for climate impact and adaptation research and (ii) unravelling inequalities in media attention to disasters in different places. Taking Germany as a case study, this paper provides an analysis of a dataset of geolocated news articles in German about landslides abroad (like those in Figure \ref{fig:news-search}), its comparison to external measures, and preliminary insights into variations in media coverage of events across countries.

\section{Related Work}
\label{sec:lit}
This work is framed by three key perspectives: efforts towards enhancing global landslide databases, news as a source of text-based disaster information, and studies of media attention to disasters.

\paragraph{Landslide databases} Landslides are inherently context-specific phenomena shaped by local conditions. Information about their occurrence is often fragmented in various local sources with limited interoperability \citep{VanDenEeckhaut2013}. As a result, studies often focus on single events (i.e.~landslides in specific regions and time) or types (e.g.~only rainfall-induced landslides). Indeed, studies often rely on national or sub-national databases \citep[\textit{inter alia}]{Mirus2020,Martha2021,Egas2024,Niyokwiringirwa2024,Bonini2025,Massey2025,Xie2026}. Fewer global databases are available to support systematic spatio-temporal analyses across regions and scales, as gathering and unifying data from various sources is a strenuous task due to their ``different criteria, description details, and temporal and spatial scales'' \citep{Gmez2023}. Besides, existing inventories can be biased toward areas with higher reporting capacity \citep{Emberson2020}.

\paragraph{Texts in climate impact and adaptation} News archives and other collections of digital documents are a means to expand landslide databases with more events or metadata \citep{DomnguezCuesta1999,Taylor2015,Kreuzer2020,Franceschini2022}. This is a particular case of a current trend in text-based climate impact research, aiming to create inventories of hazards and disasters based on information extraction from news texts \citep{sodoge_automatized_2023,henrique_lima_alencar_flash_2024,Nguyen2024,Avcolu2025,Valkenborg2026}. News can additionally be used to validate models of landslides and other hazards such as risk assessment and warning systems \citep{Battistini2017,Yagoub2020}.

\paragraph{News geolocation and analysis} The dynamics of how climate topics and events are covered in the news has been under active investigation \citep{Yan2015,cai_2025,Roth2025,Kong07022026}. We build upon those works from a new angle: how German newspapers portray landslides abroad. To geolocate news, we follow the practice of information extraction with Large Language Models (LLM) \citep{li-etal-2024-using-llms,Bauman2025,kriesch_geolocated_2025}.

\section{Data Preparation}
We derived a sample of news about landslides geolocated at the country level from the news dataset about extreme climate events and other disasters created by \citet{madureira-2026}. Details on the selection and annotation process can be found in that publication; here, only the relevant aspects are summarised. 

A collection of 317,036 news articles in German from 2000 to 2024 was retrieved from the \texttt{wiso-net} news aggregator database\footnote{Available at \url{https://www.wiso-net.de/}.} using 17 German landslide-related keywords (such as \textit{landslide}, \textit{rockfall} and \textit{mass movement}; see Appendix). After the preprocessing steps to reduce the number of unrelated or local news, 88,940 candidate documents remained. Among them, 450 unique texts were randomly sampled (uniformly over the years), and annotated by humans with a binary label for relevance (i.e.~whether the news refer to actual landslides) and the country where the landslide occurred. 58.6\% of the annotated documents were relevant.

\paragraph{Processing} We used the open-weight, locally run LLM \texttt{google/gemma-4-31b} to first identify all documents that report on specific past landslides and, then, geolocate (at the country level) the ones classified as relevant. This was performed in a two-pass procedure using the prompts displayed in the Appendix. In the first step, the model predicted a binary relevance label; in the second, it was prompted to output a JSON-formatted string with the ISO-3 code of the countries where landslides were reported.

\subsection{Evaluation on the Gold Standard}

To ensure the quality of the resulting dataset, we first assessed the LLM's performance on the human-annotated sample (i.e.~the gold standard).

\paragraph{Relevance classification} The model achieved binary precision, recall and F1-score of 0.865, 0.973 and 0.916, respectively. We manually inspected all 40 false positives. While 5 were judged as really not relevant, the remaining ones were either wrongly annotated or at least related to the overall \textit{topic} of landslides, e.g.~referring to the risk, possible impact or regularity of such events. Although we focus on landslides \textit{de facto}, the inclusion of some false positives of this type is acceptable, since established risks can already lead to consequences such as official warnings or prevention measures.

\paragraph{Geolocation} The accuracy was 0.875 for exact matches (i.e.~the extracted countries fully matched the annotation) and 0.924 for overlaps (i.e.~at least one extracted country matched the annotation). In the gold standard, 242 documents were about a single country, 19 about 2, and 3 about 3 countries. The LLM extraction followed a similar distribution: 241 single-country documents, 18 about 2 and 5 about 3. There was thus no evidence of long lists of spurious countries inflating the overlap accuracy.

\smallskip

These results indicate that the news sample had sufficient precision and recall and that the geolocation was accurate enough for subsequent analyses. 

\subsection{Independent External Datasets}

The list of countries by the United Nations Statistics Division was used as a reference. After a few needed adjustments (see Appendix), 244 countries remained. For each country, we compiled the following country-level indicators to aid the analysis and validation (download links in the Appendix): 

\begin{itemize}
	\setlength\itemsep{0em}
	
	\item \textbf{Landslide occurrence (EM-DAT)}: the number of landslide events (main or associated type) from 2000 to 2024 recorded in the International Disaster Database (2,071 in total).
	
	\item \textbf{Landslide frequency (WB-GLHM)}: the estimated average annual number of (flood and earthquake-triggered) landslides predicted by the World Bank's Global Landslide Hazard Map.

	\item \textbf{Landslide risk}: the landslide risk, categorised as high, medium, low and very low, computed by the ThinkHazard!~project of the World Bank Group and the Global Facility for Disaster Reduction and Recovery.
	
	\item \textbf{Development status}: the country's development level (developed/developing), mapped to Global North and South, respectively, as disclosed by the United Nations Conference on Trade and Development.

	\item \textbf{Income group}: the income level (low, lower middle, upper middle or high) as classified by the World Bank.
	
\end{itemize}

\section{Methods}
\begin{figure}[t]
	\centering
	\includegraphics[width=\columnwidth,trim={0 0cm 0cm 0},clip]{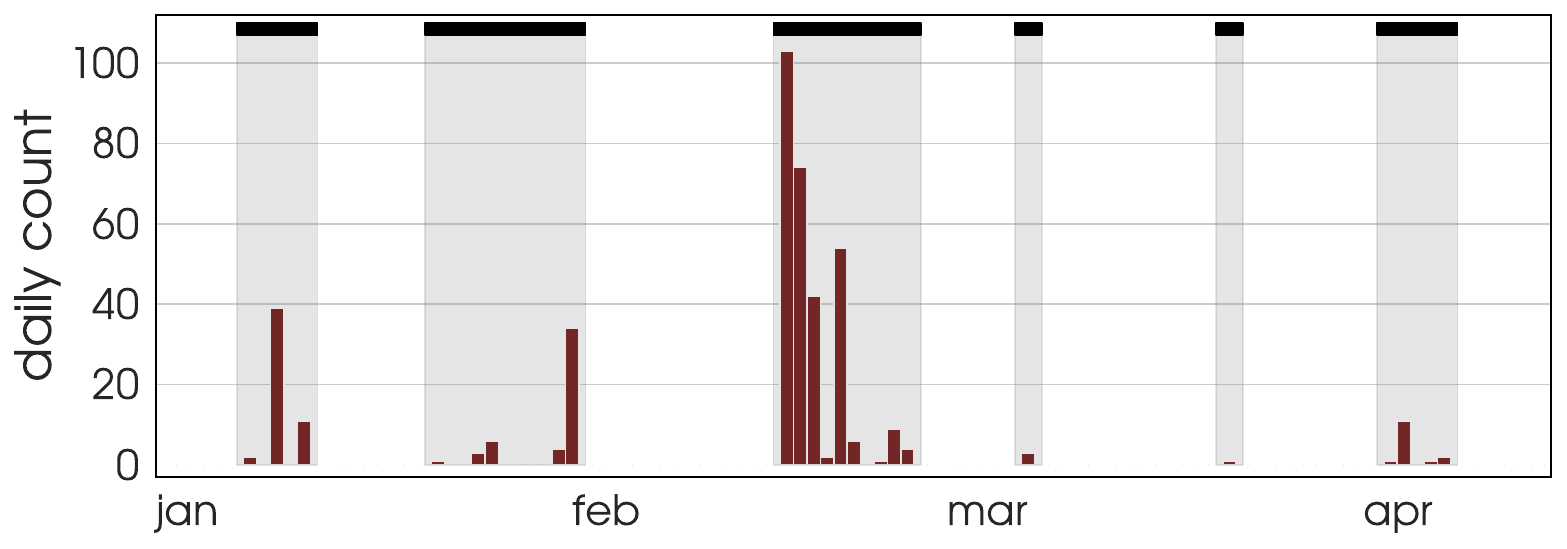}
	\caption{Example of the segmentation of a count time series into news events using the data instances about Brazil from Jan to mid-Apr, 2022. Grey-shaded areas represent each distinct identified news event.}
	\label{fig:segmentation}
\end{figure}

We created a \textit{count time series} \citep{Davis2021} for each country, i.e.~a series of non-negative integers corresponding to the daily number of news about landslides in the period from Jan 1, 2000 to Dec 31, 2024. Duplicate texts published in different news outlets were counted as distinct documents as they increase the level of media attention. Since some documents refer to landslide events in more than one country, separate \textit{data instances} were created for each mentioned country. 

 Two steps followed: (i) the identification of \textit{news events} and (ii) a comparison of the number of identified news events to external measures, in order to assess whether the distribution of landslide-related news aligned with expected landslide occurrence.

\begin{figure*}[!th]
	\centering
	{%
		\setlength{\fboxsep}{0pt}%
		\setlength{\fboxrule}{0.5pt}%
		\fbox{\includegraphics[width=0.89\textwidth,trim={0.5cm 0.6cm 0.5cm 0.6cm},clip]{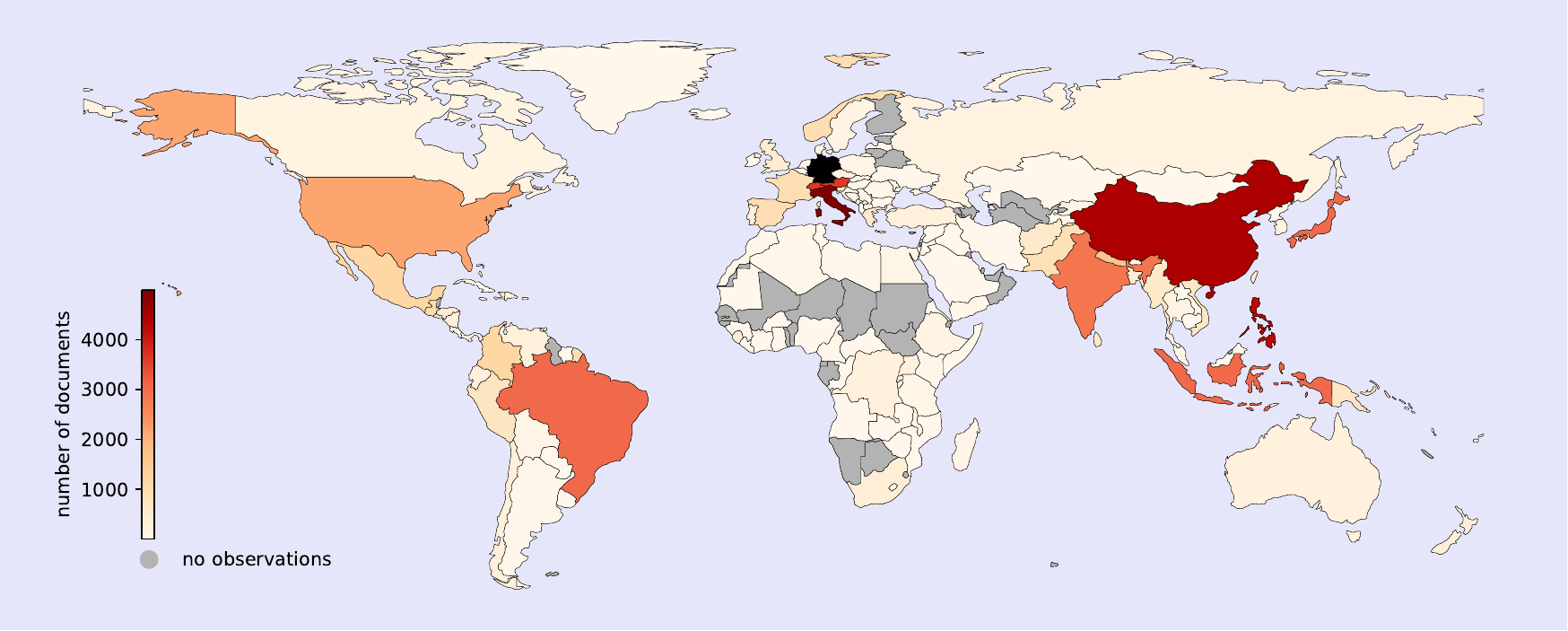}}%
	}%
	\caption{Total number of news documents about landslides in each country. Germany (in black) was not analysed.}
	\label{fig:countmap}
\end{figure*}

\begin{figure*}[th]
	\centering
	\includegraphics[width=0.9\textwidth]{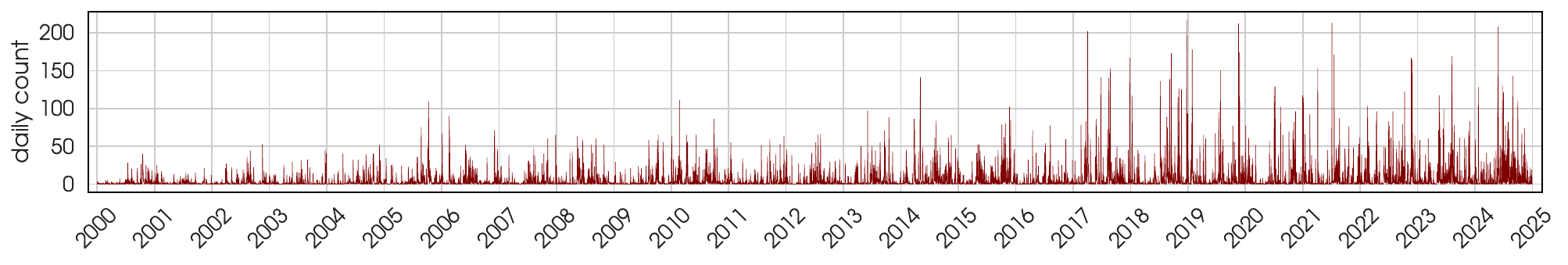}
	\caption{Overview of the count time series of daily news about landslides abroad in German newspapers. The increase in recent years may partially reflect the underlying temporal bias of the news database rather than necessarily an increase in landslide events or in media attention.}
	\label{fig:timeseries}
\end{figure*}

\subsection{Identification of News Events}

Each country's time series was segmented into \textit{news events}, defined here as temporally contiguous clusters of news discussing landslides in that country. The number of active days (i.e.~days with at least one document, representing non-zero media attention) \textit{per country} is very sparse (see Section \ref{sec:analysis}). We thus used a straightforward segmentation method: each sequence of adjacent active days was considered a distinct news event. A maximum of 4 sequential inactive days between active days was allowed during a news event. If 5 or more sequential inactive days occurred, news events were split. Figure \ref{fig:segmentation} shows an example of the segmentation. 

We measured the total number of documents, events and active days per country. For each identified news event, we computed its duration in days, its total volume of news documents and other measures listed in Table \ref{tab:measures} (Appendix).  

Note that not every news event aligns with \textit{de facto} new landslides in the real world: while many texts do discuss ongoing or latest landslides, others recall past events, refer to aggregated observations or discuss the hazard without mentioning a concrete event. The distinction is left for future work; in this study, what matters is that each news event brings the \textit{topic} of landslides in a country (back) into the media's attention. 

\subsection{Salience Scores and Divergence} 

In line with the extra and intra-media data investigation outlined by \citet{Rosengren1970}, we verified to what extent our measures aligned with independent observations. The number of identified news events per country was compared with two external measures: the number of landslide events recorded in EM-DAT for the same period and the estimated annual average in WB-GLHM. This validation step is meant for evaluating whether countries receiving more media attention are the same with higher documented landslide occurrence or susceptibility, and for investigating potential biases in news coverage.

Since no authoritative and complete global database of \textit{all} landslides exists, and news events do not always refer to new landslides, we refrained from comparing exact numbers. We were mainly interested in how many more landslides (or news events) occur or are expected in a country in comparison to other countries. For that purpose, we computed \textit{salience scores} based on the number of news events and the two external measures, by taking the log and min-max scaling the values to the interval $[0,1]$. Thus, for each measure, countries were mapped to a continuous scale ranging from 0 (least salient) to 1 (most salient). 

If newsworthiness was proportionally distributed across countries, those that score higher (lower) in landslide occurrence should also score higher (lower) in observed news events. Deviations from that proportional relation suggest that a country has been receiving more or less media attention than expected. To operationalise this rationale, we computed the residuals of a linear regression as a proxy for salience divergence, using news salience as the dependent variable and external salience as the independent variable:

{\small 
\begin{equation*}
	divergence = salience_{news} - (\hat{\beta}_0 + \hat{\beta}_1salience_{ext}) 
\end{equation*}
}

where $\hat{\beta}_{\{0,1\}}$ are estimated parameters. Therefore, the difference between the observed news salience score and the prediction by the linear regression was interpreted as a deviation from what could be expected based on external measures. 

Divergence close to 1 (-1) means that a country scores higher (lower) in news salience than in EM-DAT or WB-GLHM salience. Values near 0 indicate that news salience is close to what is expected. A similarity threshold was set using the magnitude of the most extreme observations to select a symmetric portion of 25\% of the interval around 0, considered the fitting (i.e.~non-divergent) salience range. Based on that, we grouped countries into three categories: those with news salience higher (overreported), similar or lower (underreported) compared to the external salience. 

\section{Results and Analysis}
\label{sec:analysis}	
First, we present the estimated properties of German news about international landslides. Then, we compare our measures to external sources.

\subsection{Overview: Landslides in German News}

The resulting dataset contains 54,806 documents published by 239 news outlets with 25,294 text types. Figure \ref{fig:timeseries} is an overview of the complete count time series depicting their temporal distribution by date. 5,098 active days occur in the 9,132-day period. Almost 45\% of the days have none and 16.7\% have only one document. 52 days are peaks with at least 100 documents, with maximum 217.  

In total, there are 59,997 data instances covering 152 countries (after discarding 209 occurrences of 24 spurious ISO codes generated by the LLM) whose spatial distribution is shown in Figure \ref{fig:countmap}. 92.8\% of the documents are about one country and 5.4\% about two. Documents concentrate in a few countries: the top ten (shown in Table \ref{tab:top-ten}) account for 58.5\% of the total number of instances. Only 28 countries have 500 or more instances, covering 83.1\% of the total; 92 countries have none and 45 have only up to 10 instances. Country-specific series are extremely sparse: the first, second and third quartiles for the number of active days are 3, 14 and 53.25, respectively. The most frequent country has 638 active days (still $<$7\% of all days).

\begin{table}
	\centering
	\small
	\begin{tabular}{lrrr}
		\toprule
		\textbf{country} & \textbf{news events} & \textbf{~~~~~~~docs} & \textbf{active days} \\
		\midrule

			Italy & 314 & 4,991 & 631 \\
			Switzerland & 306 & 3,640 & 638 \\
			Austria & 237 & 3,617 & 521 \\
			China & 217 & 4,438 & 580 \\
			USA & 191 & 2,172 & 356 \\
			India & 173 & 2,853 & 382 \\
			Philippines & 153 & 4,244 & 411 \\
			Nepal & 142 & 1,662 & 312 \\
			Brazil & 138 & 3,056 & 356 \\
			Indonesia & 136 & 3,059 & 291 \\
			France & 120 & 953 & 175 \\
			Spain & 116 & 974 & 173 \\
			Japan & 114 & 3,045 & 275 \\
			Peru & 103 & 733 & 166 \\
			Colombia & 100 & 1,161 & 150 \\
			Pakistan & 89 & 890 & 172 \\
			Norway & 80 & 962 & 118 \\
			Czechia & 74 & 158 & 96 \\
			UK/N.~Ireland & 67 & 443 & 103 \\
			Mexico & 61 & 1,170 & 138 \\

		\cline{1-4}
		\bottomrule
	\end{tabular}
	\caption{The 20 countries with the most identified news events and the corresponding number of documents and active days in the period. Full table in the Appendix.}
	\label{tab:top-ten}
\end{table}

Figure \ref{fig:outlets} shows the number of documents for the most frequent outlets and their distribution per continent. Documents about Africa and Oceania occur less often across all news outlets. Asia is the most represented continent, followed by Europe and the Americas, with a few exceptions.

\begin{figure}[t]
	\vspace{-0.3cm}
	\includegraphics[width=\columnwidth]{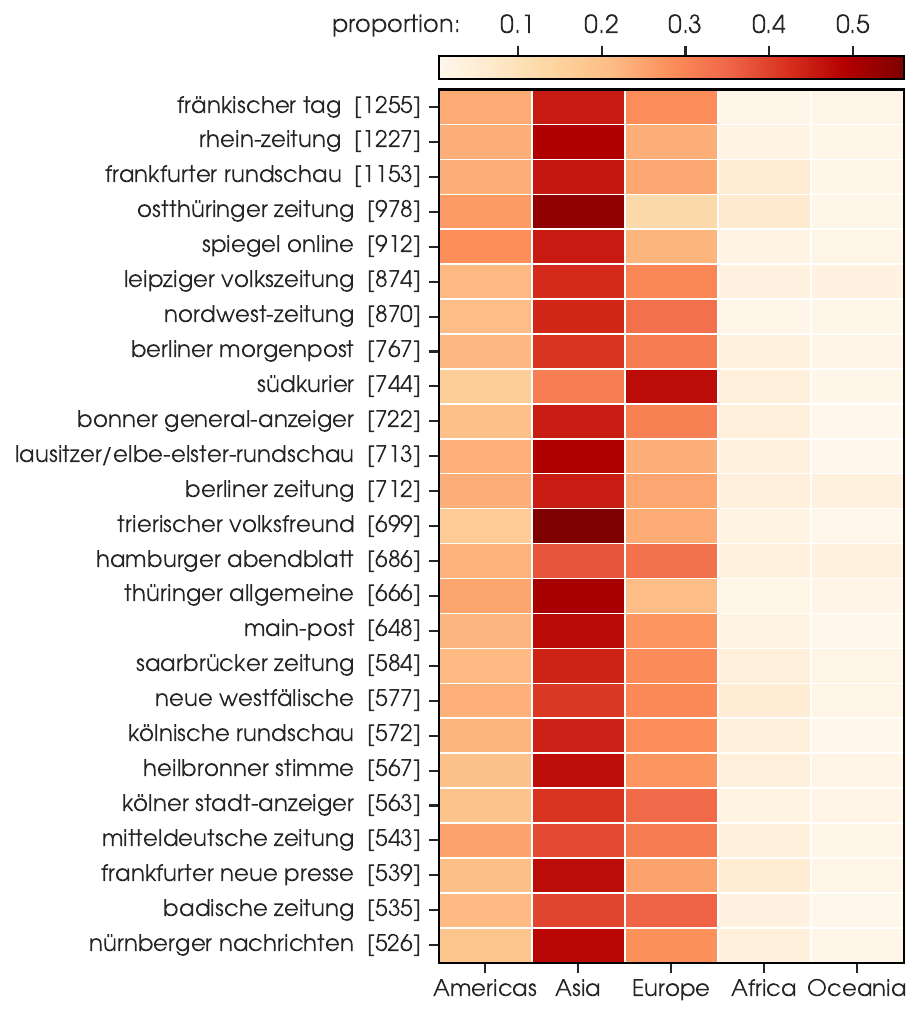}
	\caption{Number of documents for each news outlet (>500) and distribution of data instances by continent.}
	\label{fig:outlets}
\end{figure}

\begin{figure*}[t]
	\centering

	\begin{subfigure}{0.49\textwidth}
		\includegraphics[width=\textwidth]{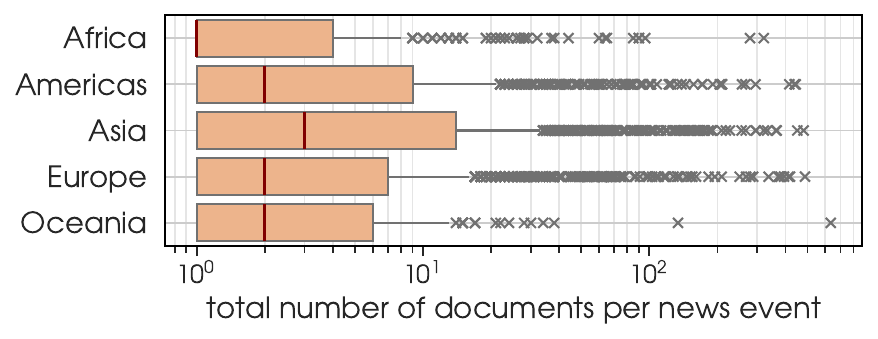}
		\caption{Log-plot of the distribution of volume by continent.}
		\label{fig:volume}
	\end{subfigure}
	\hfill
	\begin{subfigure}{0.49\textwidth}
		\includegraphics[width=\textwidth]{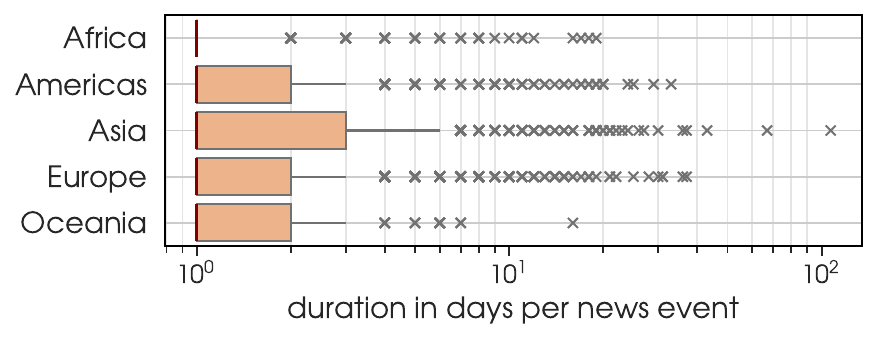}
		\caption{Log-plot of the distribution of duration by continent.}
		\label{fig:duration}
	\end{subfigure}

	\vspace{0.5cm}

	\begin{subfigure}{0.49\textwidth}
	\includegraphics[width=\textwidth]{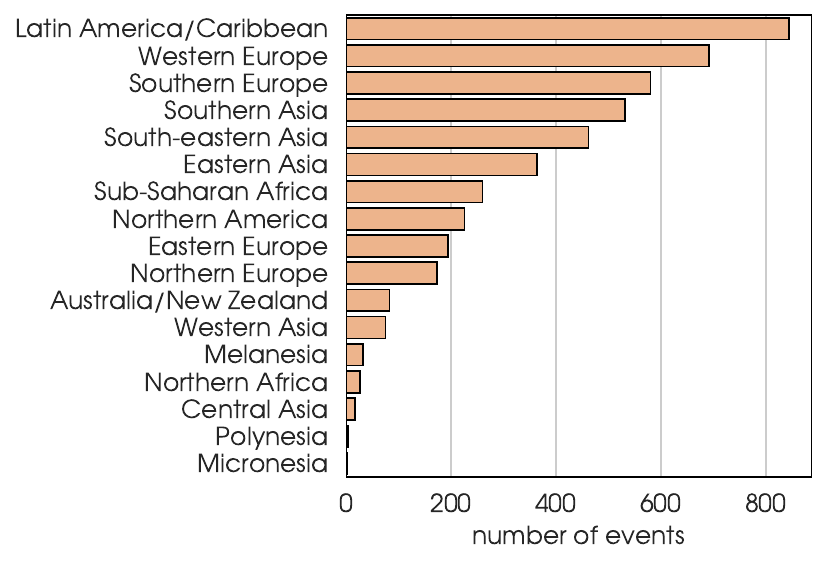}
	\caption{Distribution of identified news events by subregion.}
	\label{fig:events-subregion}
\end{subfigure}
\hfill
\begin{subfigure}{0.49\textwidth}
	\includegraphics[width=\textwidth]{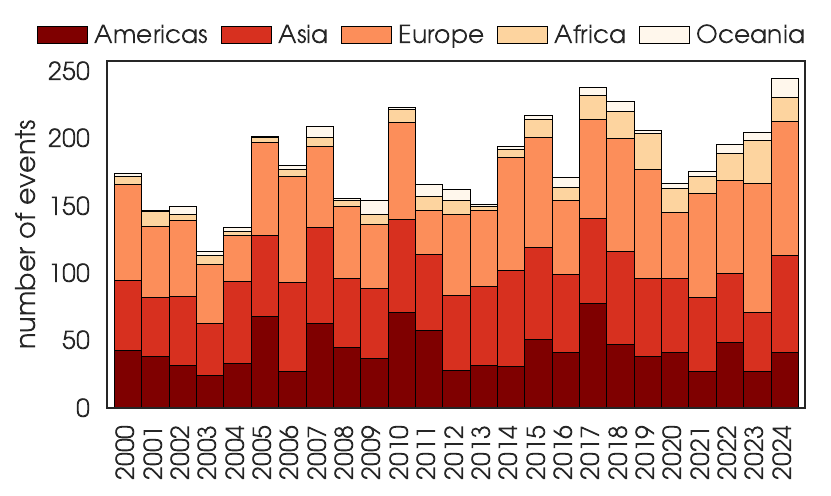}
	
	\vspace{0.7cm}
	\caption{Distribution of identified news events by year and continent.}
	\label{fig:events-year}
\end{subfigure}

	\caption{Overview of identified news events.}
	\label{fig:figures}
\end{figure*}

\subsection{Spatio-Temporal Distribution of News Events} 

The segmentation of the geolocated news reporting on landslides led to the identification of 4,567 news events. Countries with the largest number of news events are listed in Table \ref{tab:top-ten}. The rankings for number of events, documents and active days are not the same for all countries, making these three measures complementary. Some countries have more documents forming fewer news events (e.g.~China) or more news events with fewer documents (e.g.~USA); similarly, some countries have more documents spread across fewer days (e.g.~Indonesia) and vice versa (e.g.~France).

The duration and volume measures help capture such variations. The log-plots in Figures \ref{fig:volume} and \ref{fig:duration} show that the median number of days per news event is 1 for all continents and the median number of documents is between 1 and 3. Most news events are brief (1-3 days) with no more than 14 documents, but several outliers capture more media attention. Figures \ref{fig:events-subregion} and \ref{fig:events-year} show the spatial and temporal distribution of identified news events. Despite some yearly variation, there is no clear bias towards recent years as in the daily counts.

\subsection{Comparison to External Sources}

We investigate the extent to which the non-uniform distribution of news events over countries and the resulting salience scores are due to media bias or to the real occurrence of more landslides.

\paragraph{Distribution of total news events} In Table \ref{tab:valid}, the percentage of news events assigned to each country category is compared with the corresponding percentages in EM-DAT and WB-GLHM. In terms of hazard risk, our observations are as expected: most news events refer to countries classified as high risk, and the values are comparable to the external sources. However, the proportion of news events for the Global North is higher compared to EM-DAT and WB-GLHM, which concentrate in the Global South. This relates to continents: news events in Europe are much more frequent than in the external sources, whereas Asia is less frequent. Table \ref{tab:valid-subregion} in the Appendix shows the values by subregion, particularly indicating the overreporting of Southern and Western Europe and underreporting of South-Eastern Asia. There is also a contrast in relation to income: news events concentrate more on high income countries. This difference comes at the cost of fewer observations than expected in the upper and lower middle income, but not so much in low income countries.

\begin{table}[t]
	\centering
	\small
	\addtolength{\tabcolsep}{-0.2em}
	\begin{tabular}{ll|rr|rr}
		\toprule
		&  &  \multicolumn{2}{c|}{\textbf{news}}  & \textbf{\% EM- }& \textbf{\% WB-}\\
		&  & \textbf{$n$} & \textbf{\%} & \textbf{DAT} & \textbf{GLHM}\\
		
		\midrule
		\multirow[t]{5}{*}{\textbf{risk}} & high & 4,224 & 92.49 & 93.61 & 97.86 \\	
		& medium & 217 & 4.75 & 3.29 & 1.14 \\
		& unknown & 1 & 0.02 & 0.00 & 0.08 \\
		& low & 89 & 1.95 & 2.95 & 0.75 \\
		& very low & 36 & 0.79 & 0.15 & 0.17 \\

		\cmidrule{1-6}
		\multirow[t]{2}{*}{\textbf{develop.}} & global N. & 2,084 & 45.63 & 13.88 & 24.41 \\
		& global S. & 2,483 & 54.37 & 86.12 & 75.59 \\		
		
		\cmidrule{1-6}
		\multirow[t]{5}{*}{\textbf{continent}} & Africa & 287 & 6.28 & 9.39 & 4.16 \\
		& Americas & 1,070 & 23.43 & 27.14 & 25.73 \\
		& Asia & 1,450 & 31.75 & 53.12 & 57.54 \\
		& Europe & 1,639 & 35.89 & 7.55 & 7.96 \\
		& Oceania & 121 & 2.65 & 2.81 & 4.60 \\
		
		\cmidrule{1-6}
		\multirow[t]{5}{*}{\textbf{income}} & high  & 2,183 & 47.80 & 16.98 & 28.02 \\
		& upper mid & 1,129 & 24.72 & 39.86 & 36.91 \\
		& unknown & 48 & 1.05 & 1.31 & 1.13 \\
		& lower mid & 1,002 & 21.94 & 33.62 & 30.08 \\
		& low & 205 & 4.49 & 8.22 & 3.86 \\
		\cline{1-6}
		
		\bottomrule
	\end{tabular}
	\caption{Number and percentage of identified news events compared with external measures.}
	\label{tab:valid}
\end{table}

\paragraph{Temporal aspect} We examine the distribution of news events by subregion for each year and compare the observed proportions with their external counterparts. The distribution of differences is displayed in Figure \ref{fig:years-box-wb} for WB-GLHM (the equivalent plot for EM-DAT is in the Appendix). The yearly proportions generally deviate by only up to 5\% in relation to the external measures. Noticeable deviations occur for South-eastern Asia and Northern America, and partially Eastern Asia, with lower yearly proportions, and for Southern and Western Europe, as well as partially Latin America/Caribbean, with larger yearly proportions. The comparison to EM-DAT is mostly similar, except for Northern America, which is within the 5\% range, and for opposite results for Latin America/Caribbean. Detailed plots by year are in Figure \ref{fig:years-subregions} (Appendix).

\begin{figure*}[th]
	\begin{subfigure}[t]{0.68\textwidth}
		{%
			\setlength{\fboxsep}{0pt}%
			\setlength{\fboxrule}{0.5pt}%
			\fbox{\includegraphics[width=\textwidth,trim={0.5cm 0.3cm 0.5cm 0.5cm},clip]{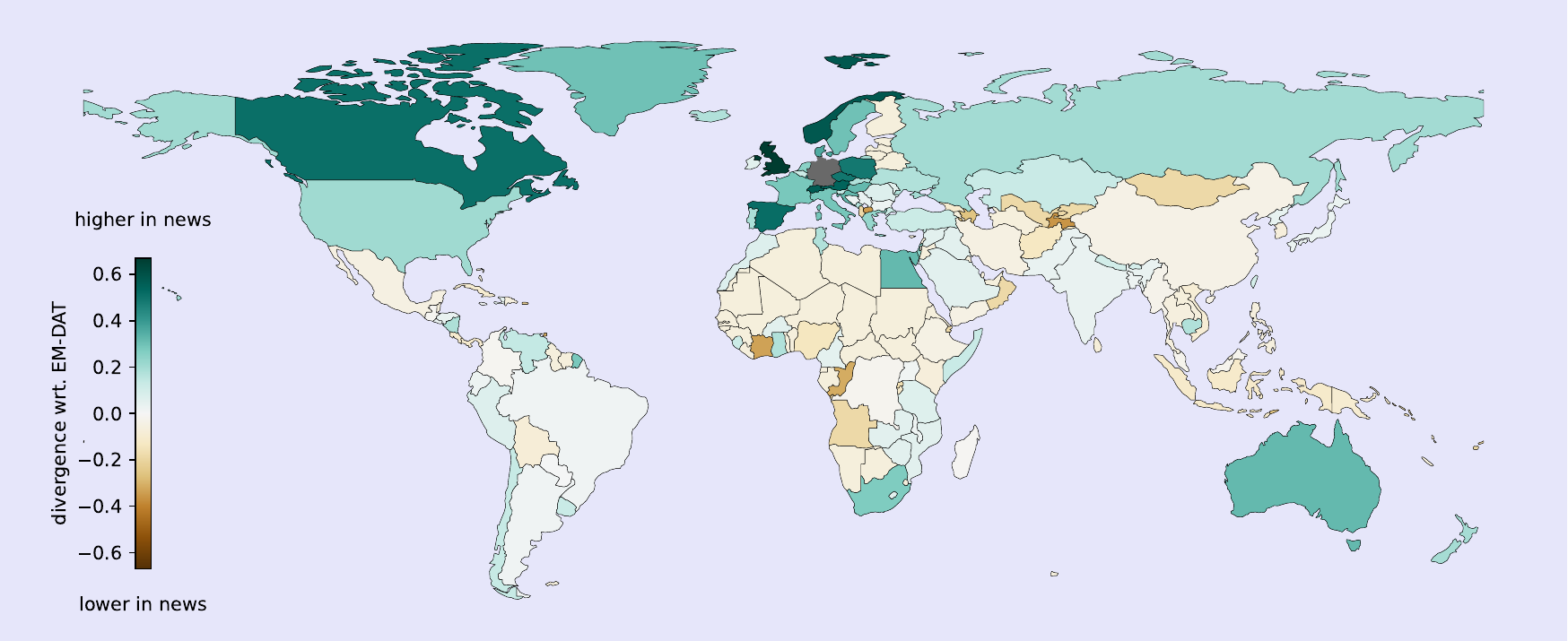}}%
		}
		\caption{Divergence scores: EM-DAT vs.~news events}
		\label{fig:scores-emdat}
	\end{subfigure}
	\hfill
	\begin{subfigure}[t]{0.3\textwidth}
		\includegraphics[width=\textwidth]{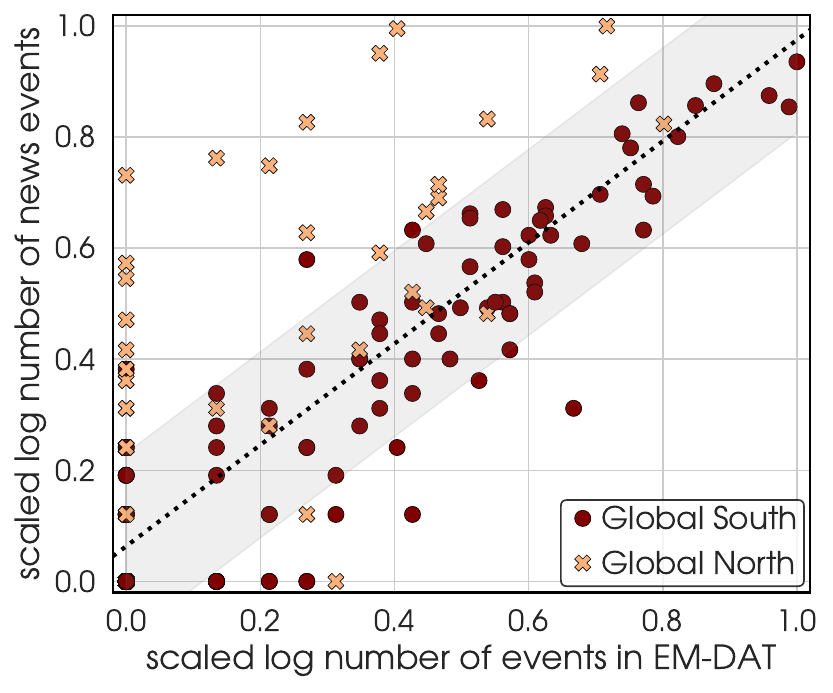}
		\caption{Salience: EM-DAT vs.~news.}
		\label{fig:scatter-emdat}
	\end{subfigure}%
	\vspace{0.3cm}
	\begin{subfigure}[t]{0.68\textwidth}
		{%
			\setlength{\fboxsep}{0pt}%
			\setlength{\fboxrule}{0.5pt}%
			\fbox{\includegraphics[width=\textwidth,trim={0.5cm 0.3cm 0.5cm 0.5cm},clip]{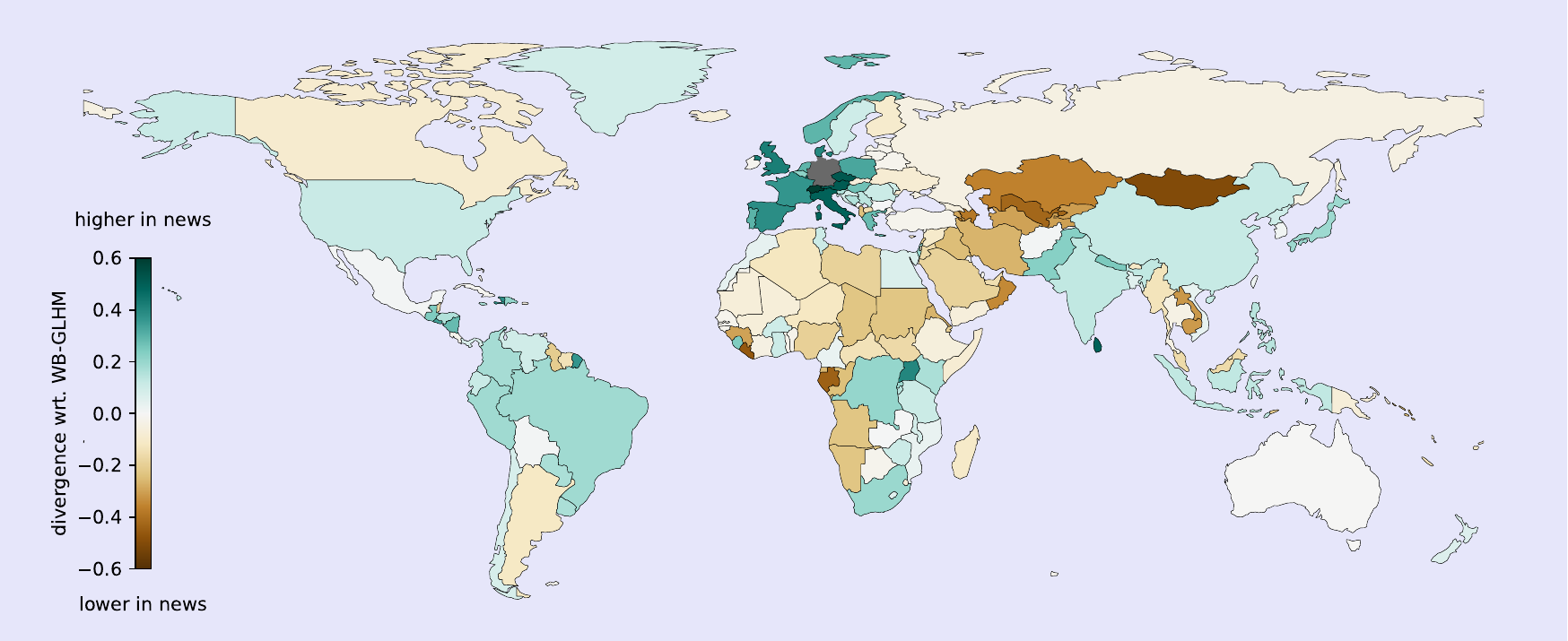}}%
		}
		\caption{Divergence scores: WB-GLHM vs.~news events.}
		\label{fig:scores-wb}
	\end{subfigure}
	\hfill
	\begin{subfigure}[t]{0.3\textwidth}
		\includegraphics[width=\textwidth]{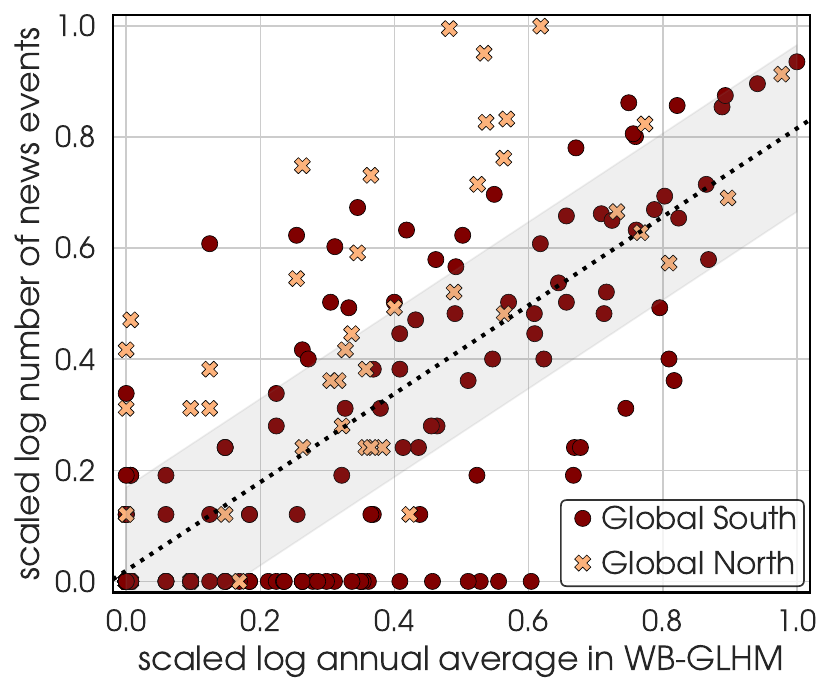}
		\caption{Salience: WB-GLHM vs.~news.}
		\label{fig:scatter-wb}
	\end{subfigure}
	
	\caption{Divergence scores and linear regression with the defined similarity boundary (in grey) for salience scores.}
	\label{fig:scores}
\end{figure*}

\paragraph{Salience and divergence scores} The divergence scores for all countries are displayed in Figures \ref{fig:scores-emdat} and \ref{fig:scores-wb}, contrasting negative and positive divergence. The scatter plots in Figures \ref{fig:scatter-emdat} and \ref{fig:scatter-wb} help visualise how salience scores are distributed around the fitted line. For EM-DAT, the line coefficient and intercept are 0.91 and 0.06, respectively; for WB-GLHM, the estimates are 0.79 and 0.01, respectively. The grey area represents the computed similarity range. Data points above (below) it are countries considered to be overreported (underreported) in German news. Exact values by country are presented in the Table \ref{tab:full-scores} (Appendix). 

The percentage of observations in each divergence group is displayed in Table \ref{tab:score-diffs} by development level. Most countries fall within similar ranges on both external measures, corroborating the overall validity of the dataset. Still, some countries both in the Global North and South are more/less salient in the news than expected. A considerable portion of countries in the Global North are more salient in the news than expected and only two are underreported (Albania and North Macedonia) based on both external measures, whereas the portion of countries in the Global South that are less salient than expected is higher for WB-GLHM than EM-DAT. 

The most extreme underreporting (negative divergence) with respect to EM-DAT occurs for Tajikistan, North Macedonia, Côte d’Ivoire, Trinidad and Tobago, and Congo; the overreporting (positive divergence) extremes are Spain, Austria, Switzerland, Norway and UK/Northern Ireland. For WB-GLHM, the negative extremes include Mongolia, Liberia, Gabon, Uzbekistan and Azerbaijan; the positive extremes are Italy, Sri Lanka, Austria, Czechia and Switzerland. Only two countries fall into opposite extremes between EM-DAT and WB-GLHM: Cambodia (higher for EM-DAT but lower for WB-GLHM) and Rwanda (lower for EM-DAT but higher for WB-GLHM).  

\begin{figure}[t]
	\includegraphics[width=\columnwidth]{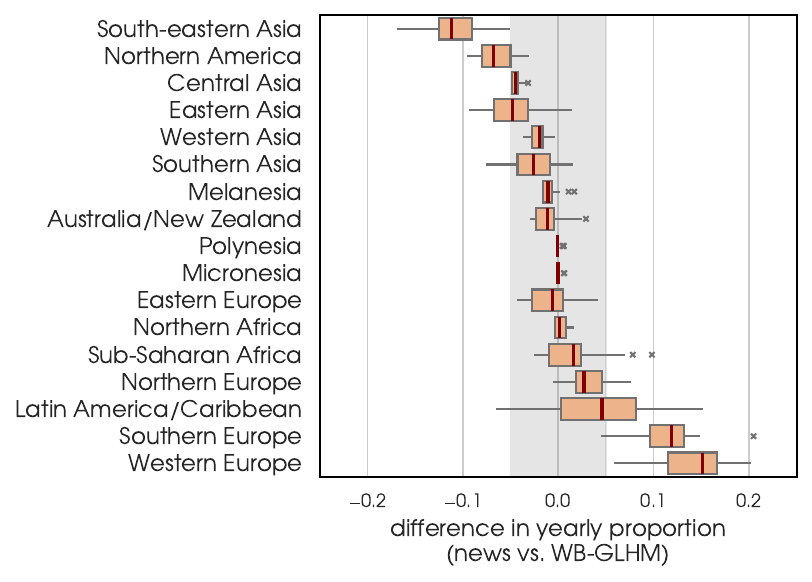}
	\caption{Box-plot of the differences in yearly proportion between news events and WB-GLHM.}
	\label{fig:years-box-wb}
\end{figure}

\begin{table}[t]
	\centering
	\small
	\addtolength{\tabcolsep}{-0.35em}
	\begin{tabular}{p{1.7cm} m{1.8cm} m{1cm} m{1cm} m{1cm}}
		\toprule
		& & \textbf{lower in news} & \textbf{similar scores} & \textbf{higher in news} \\
		\midrule
		news vs. & Global North & ~~3.12 & 53.12 & 43.75 \\
		EM-DAT & Global South & 11.67 & 85.00 & ~~3.33 \\
		\cmidrule{1-5}
		news vs. & Global North & ~~3.12 & 65.62 & 31.25 \\
		WB-GLHM & Global South & 21.67 & 65.56 & 12.78 \\
		\cline{1-5}
		\bottomrule
	\end{tabular}
	
	\caption{\% of under/overreported countries.}
	\label{tab:score-diffs}
\end{table}

\section{Discussion}
According to our estimations, German newspapers discussed landslides abroad more than 4.5k times in 2000-2024. Although a few countries were more salient than others in the absolute number of documents and news events, part of this imbalance followed expected patterns of susceptibility to landslides, as evidenced by historical data of observed events (EM-DAT) and a data-driven predictive model (WB-GLHM). We have validated the resulting dataset in terms of overall distribution by a country-level indicator of hazard risk and by salience scores, with some interesting deviations.

Some countries/subregions turned out to be disproportionally salient in German news: there were more news events about countries in the Global North, mostly in some regions of Europe and/or of those of high income, than expected if newsworthiness was proportionally distributed based on countries' susceptibility to landslides. Conversely, a few countries have received less attention than expected, including parts of Asia and countries with middle income levels. 

These asymmetries shed light on systematic biases in disaster reporting that demand further investigation. The particular media bias explored here suggests that German-speaking audiences receive a partially distorted picture of global landslide risk. A skewed coverage has implications for climate awareness and preparedness: it may limit public understanding of the true scale and geographic distribution of disasters, potentially reducing political will and public support for international climate adaptation financing and risk reduction initiatives. 

The different results in some of the comparisons to EM-DAT and WB-GLHM ought to derive from their distinct nature: EM-DAT only considers observed events that made their way to a few curated sources (such as selected press agencies, governmental agencies and insurance companies) and surpassed a given impact level. Due to geographical and cultural proximity, German newspapers possibly often report on local, less severe events in Europe that do not meet such criteria (e.g.~roads blocked by landslides), which could explain the higher proportion of news events in Europe. The WB-GLHM model, on the other hand, is based on real-world environmental observations. It should account for all types of events, despite their severity; still, it is merely \textit{predictive} and may misestimate actual events in each year. Canada, for example, is the 12th most salient country in WB-GLHM but had only 1 entry in EM-DAT, leading to opposite signs in its divergence scores.

\section{Conclusions and Future Work}
We analysed a large-scale dataset of German news articles about landslides worldwide which can enrich incomplete disaster databases and enable large-N studies of inequalities in media attention. Our results revealed preliminary insights regarding imbalance in coverage relative to a global landslide inventory (EM-DAT) and a predictive landslide frequency model (WB-GLHM). Overall, news in German about landslides tend to favour some countries and regions while overlooking others. It stands as a foundational layer towards understanding how social inequalities are reflected and reinforced by disaster coverage in Global North's media systems. 

This work paves the way for a variety of relevant research questions. Possible next steps are modelling media coverage based on countries' socio-economic, cultural, linguistic and geographical proximity to Germany; assessing the temporal
dynamics of media attention cycles; distinguishing different types of news events (i.e.~reporting about past, ongoing or future/hypothetical events, as in \citealp{huang-etal-2016-distinguishing}); employing alignment methods and information extraction techniques to extract event metadata; expanding the analysis to other hazards such as wildfires and floods; and examining the language use for reporting on landslides.

\section*{Limitations}
The LLM performance did not generalise to other types of hazards like cold waves and droughts. The prompts should thus not be directly used for other datasets without careful evaluation. Although the relevance classification and geolocation performance were high, some errors still occur, making unrelated and/or wrongly geolocated documents form spurious news events. False positives can be a problem especially for countries that are rarely observed and have few news events. Moreover, LLMs can introduce biases in geocoding, which, in turn, influence results and conclusions. A stratified analysis of the LLM’s geolocation codes has not been conducted yet.

We did not try to distinguish texts that only \textit{mention} a landlside among other topics from texts whose main purpose is to report on a landslide, such as the ones in Figure \ref{fig:news-search}. Depending on the purpose of the study using this dataset, this distinction may be required.

We inherited some known and unknown biases from the \texttt{wiso-net} database. We assumed it is representative of German newspapers, but their news extraction pipeline may not have included all relevant news documents and/or news outlets. Some documents are actually a concatenation of many pieces of news, without clear distinctive boundaries, which may negatively impact classification and geolocation. 

The minimum number of inactive days between different news events is a hyperparameter. Other values apart from 5 can be tested. Lower values lead to more news events, possibly splitting the same topic into more than one event, whereas higher values may cause different events to be merged into the same news event. 

We did not treat the temporal bias toward recent years in the number of documents here because it is not so prominent after event segmentation and because we mainly focused on the total number of events in the period. Still, depending on the type of analysis, normalisation methods for the number of documents may be necessary. 

The spatial analysis was performed only at the country level; we neither located events regionally nor normalised counts by area. Refining this aspect can be particularly relevant for large countries like Brazil, US and China. 

The validation with external data sources was performed using the EM-DAT inventory, a well-established source in the field (despite its known limitations and biases) that covers the full period of our observations and can easily be extended to other hazards in our pipeline. Future work can integrate and compare results using the Unified Global Landslide Database by \citet{Gmez2023} (although with observations only up to 2020) and the the Global Fatal Landslide Database by \citet{nhess-18-2161-2018} (only from 2004 to 2016).

\section*{Data and Code}
We do not have permission to share the text data; it must be accessed via the wiso-net database at \url{https://www.wiso-net.de/}. Details about the specific used instances are available upon request. The code used to produce the results presented in this paper is available at \url{https://codeberg.org/briemadu/landslides-german-news}.

\section*{Acknowledgments}
We thank the anonymous reviewers for their comments and suggestions.

\bibliography{bibs/custom}

\clearpage
\appendix

\section{Appendix}
\label{sec:appendix}
\subsection{Processing Details}

The list of German keywords used to query news documents in shown in Table \ref{tab:keywords}. 

The LLM was deployed via LM Studio.\footnote{\url{https://lmstudio.ai/}} The system prompt for classification was:
\begin{lstlisting}[breaklines,basicstyle=\small]
You are an expert in reading news articles carefully and classifying them very precisely based on their references to $hazard events.
\end{lstlisting}

The main prompt for classification was:
\begin{lstlisting}[breaklines,basicstyle=\small]
The German news article below must be classified with one of these labels:

- Label 1: At least one sentence in the text refers to a $hazard event.
- Label 0: The text does not refer to any $hazard event.

Some keywords for $hazard in German are: $keywords.

Only concrete, specific and natural $hazard events that happenned or are happenning in the real world are relevant. Articles that mention only the possibility or risk of such an event and general discussions about $hazardplural with no reference to an specific event must be ignored.

Decide which the correct label is and begin your answer with 0 or 1 accordingly. Do not begin your answer with any other word or symbol.

Here is the news article:

$text
\end{lstlisting}

The system prompt for geolocation was:
\begin{lstlisting}[breaklines,basicstyle=\small]
You are a very precise information extractor that identifies the country where $hazard occurred as reported in news articles.
You always output the results in JSON format and identify affected countries as ISO 3166-1 alpha-3 codes.
Example: {"country_codes": "DEU;BRA"}
\end{lstlisting}

\begin{table}[h]
	\centering
	\begin{tabular}{ll}
		\toprule
		\textbf{German} & \textbf{English} \\
		\midrule
		Erdrutsch 			 	& Landslide 		\\
		Felssturz, Felsstürz 	& Rockfall 			\\
		Schlammlawine 		 	& Mudslide			\\
		Massenbewegung 		 	& Mass movement 	\\
		Hangrutsch 			 	& Slope slide 		\\
		Hangbewegung 		 	& Slope movement   	\\
		Rutschung 			 	& Slip				\\
		Bodenrutsch 		 	& Ground slide		\\
		Hangabrutschung 	 	& Hillside slumping	\\
		Murgang 			 	& Debris flow		\\
		Gerölllawine 		 	& Debris avalanche	\\
		Rutschhang, Rutschhäng 	& Sliding slope		\\
		Rutschgefahr 			& Slip hazard		\\
		Felslawine 				& Rock avalanche	\\
		Mure 					& Mudflow			\\
		\cline{1-2}
		\bottomrule
	\end{tabular}
	\caption{German keywords used to query the news database and their (approximate) translations.}
	\label{tab:keywords}
\end{table}

The main prompt for geolocation was:
\begin{lstlisting}[breaklines,basicstyle=\small]
Analyze the news article below and identify the mentioned country or countries where $hazard occurred.
Follow these constraints:
1. Only past or ongoing $hazard are relevant. Ignore possible, future, hypothetical or fictional $hazard and all other types of events.
2. Only include countries that are stated in this news article. Do not use outside knowledge.
3. Return the result strictly in JSON format and convert country names to ISO 3166-1 alpha-3 codes.
4. If no countries are mentioned as being affected by $hazard, return N/A for the country_codes value.
5. If no past or ongoing $hazard are described, return N/A for the country_codes value.
6. Do not include any country for which only other events apart from $hazard are discussed in the text.

Here is the news article:
$text
\end{lstlisting}

\begingroup
\renewcommand{\arraystretch}{1.5}
\begin{table*}[h]
	\small
	\centering
	\begin{tabular}{p{2.5cm}p{12cm}}
		\toprule
		UNSD countries		& \url{https://unstats.un.org/unsd/methodology/m49/overview/} \\
		EM-DAT 		&  \url{https://www.emdat.be/} 		\\
		WB-GLHM		& 	\url{https://datacatalog.worldbank.org/search/dataset/0037584/global-landslide-hazard-map} \\
		TH-Risk	& \url{https://datacatalog.worldbank.org/search/dataset/0060151/thinkhazard-hazard-ranking} \\
		Development status	& \url{https://unctadstat.unctad.org/EN/Classifications.html} \\	
		Income group		& \url{https://datahelpdesk.worldbank.org/knowledgebase/articles/906519-world-bank-country-and-lending-groups} \\
		Polygons		& \url{http//www.naturalearthdata.com/download/110m/cultural/ne_110m_admin_0_countries.zip} \\
		\cline{1-2}
		\bottomrule
	\end{tabular}
	\caption{Download links for the external data sources.}
	\label{tab:data-links}
	\vspace{1cm}
\end{table*}
\endgroup

\begin{table*}
	\centering
	\begin{tabular}{ll}
		\toprule
		\textbf{measure} & \textbf{description} \\
		\midrule
		$n$ at peak & daily article count at the event's peak date \\
		total volume & total number of articles during the news event \\
		duration & total number of days in the news event \\
		days since last & number of days since the last news event \\
		days to peak & number of days from the first article's date until the peak date \\
		days to fade & number of days from the peak date to the last article's date \\
		$n$ text types & number of unique text types published during the news event \\
		$n$ outlets & number of media outlets reporting during the news event \\
		\cline{1-2}
		\bottomrule
	\end{tabular}
	\caption{List of all measures computed for the identified news events, based on \citep{madureira-br-2026}.}
	\label{tab:measures}
	\vspace{1cm}
\end{table*}

\paragraph{Countries} Data instances about Germany were ignored, as we are only interested in events in other countries. Taiwan was added to the reference list of countries as it appears in the predictions and in country indicators. Antarctica, Åland Islands, United States Minor Outlying Islands and Niue were excluded due to lack of data. The countries' polygon coordinates were retrieved from the Natural Earth Data project. The download links for all country indicators are listed in Table \ref{tab:data-links}.

For computing the distributions in Figure \ref{fig:outlets}, documents about more than one continent were counted as separate instances.

\subsection{Additional Results}

\begin{figure*}
	\centering
	{%
		\setlength{\fboxsep}{0pt}%
		\setlength{\fboxrule}{0.5pt}%
		\fbox{\includegraphics[width=0.89\textwidth,trim={0.5cm 0.6cm 0.5cm 0.6cm},clip]{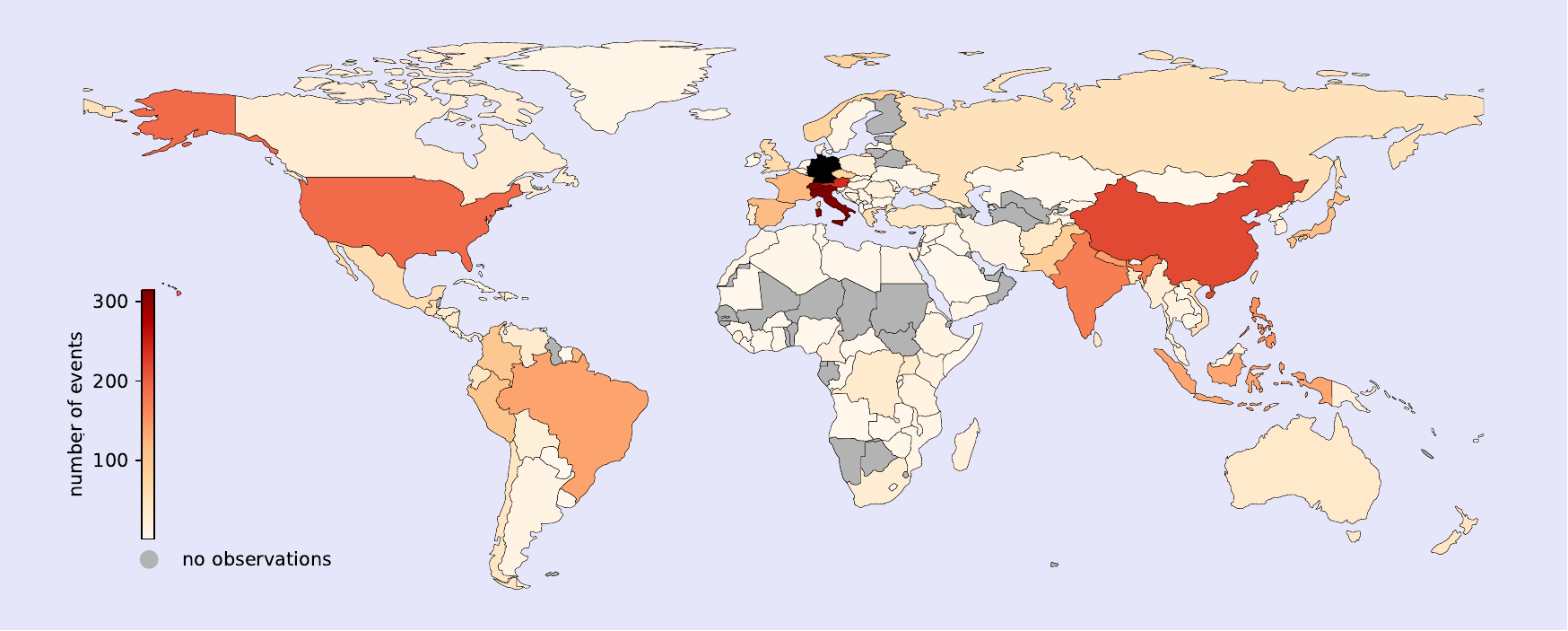}}%
	}%
	\caption{Total number of news events about landslides in each country. Germany (in black) was not analysed.}
	\label{fig:news-events}
\end{figure*}

\begin{figure*}
	\centering
	{%
		\setlength{\fboxsep}{0pt}%
		\setlength{\fboxrule}{0.5pt}%
		\fbox{\includegraphics[width=0.89\textwidth,trim={0.5cm 0.6cm 0.5cm 0.6cm},clip]{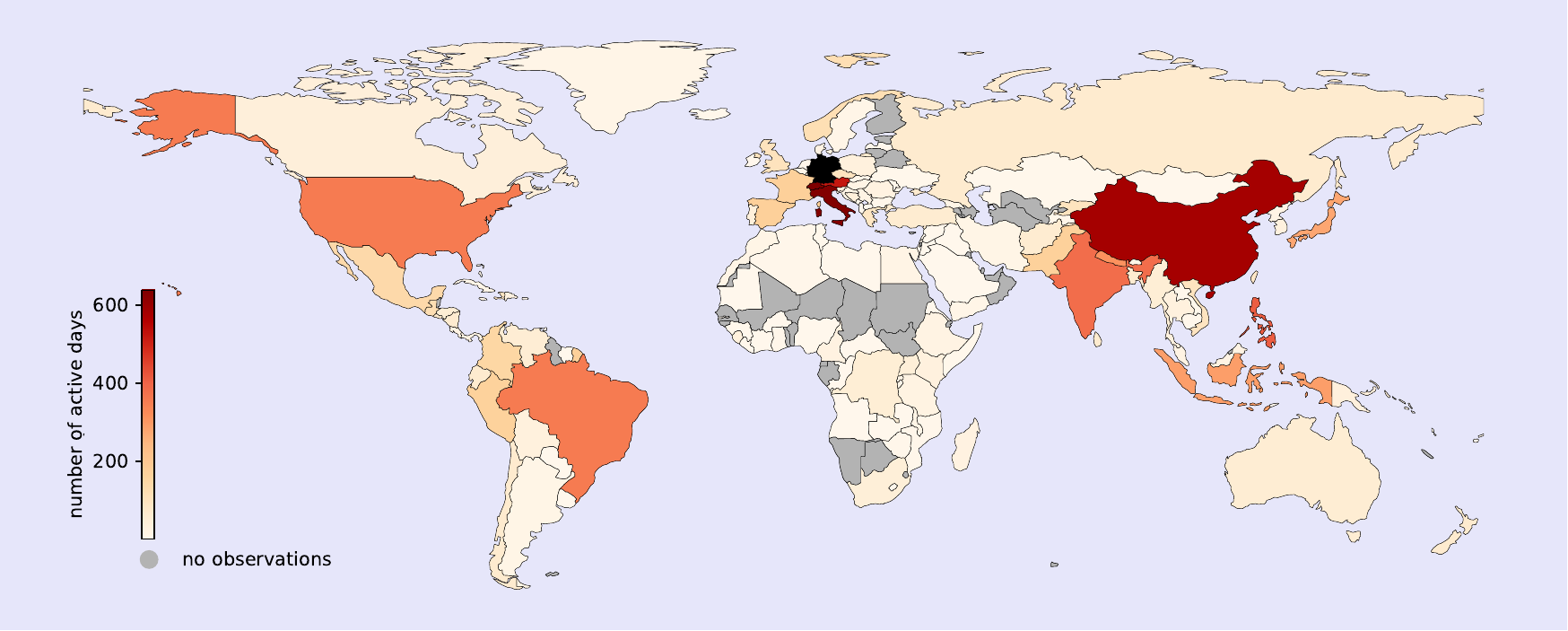}}%
	}%
	\caption{Total number of active days for each country. Germany (in black) was not analysed.}
	\label{fig:active-days}
\end{figure*}

\begin{figure}
	\centering
	\includegraphics[width=\columnwidth]{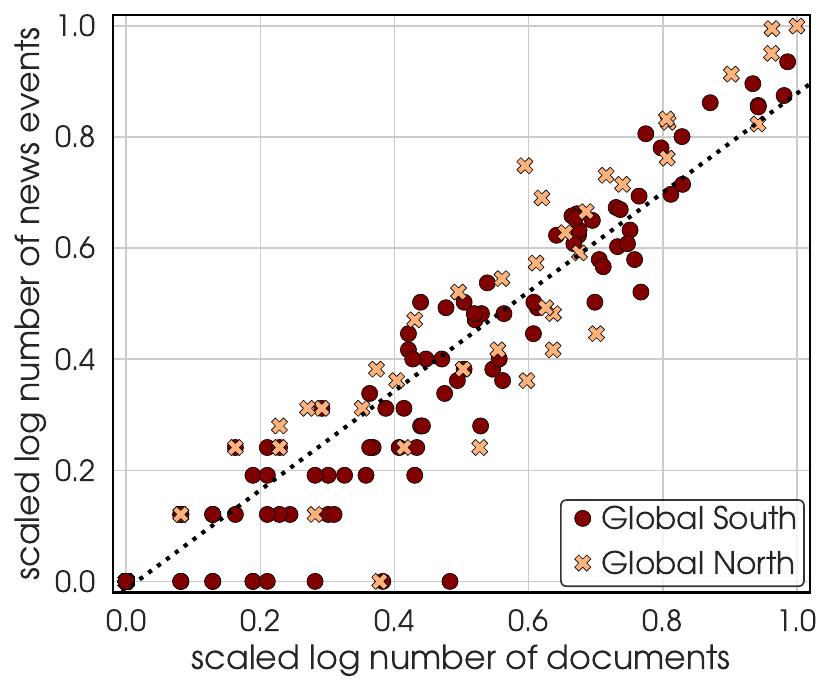}
	\caption{Relation between number of documents and number of news events.}
	\label{fig:scatter-docs}
\end{figure}

\begin{figure}
	\centering
	\includegraphics[width=\columnwidth]{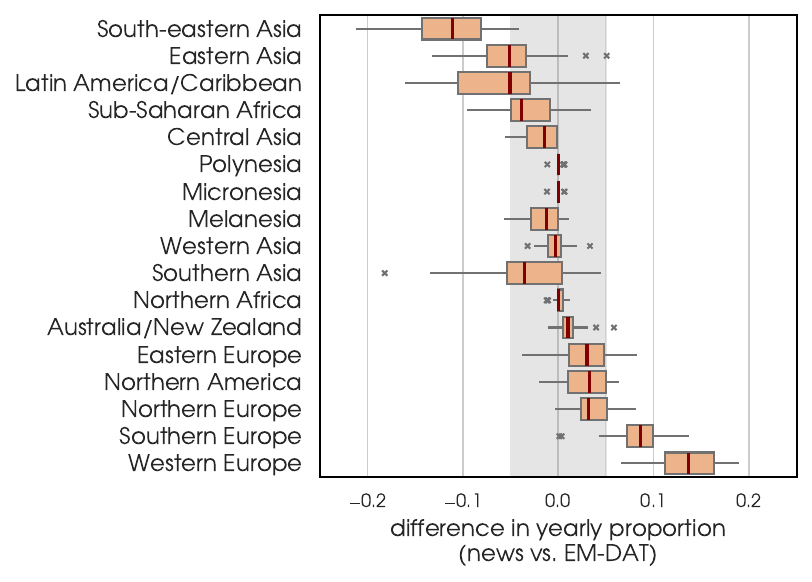}
	\vspace{0.2cm}
	\caption{Box-plot of the differences in yearly proportion of news events and EM-DAT.}
	\label{fig:years-box-emdat}
\end{figure}

\begin{figure*}[h]
	\includegraphics[width=\textwidth]{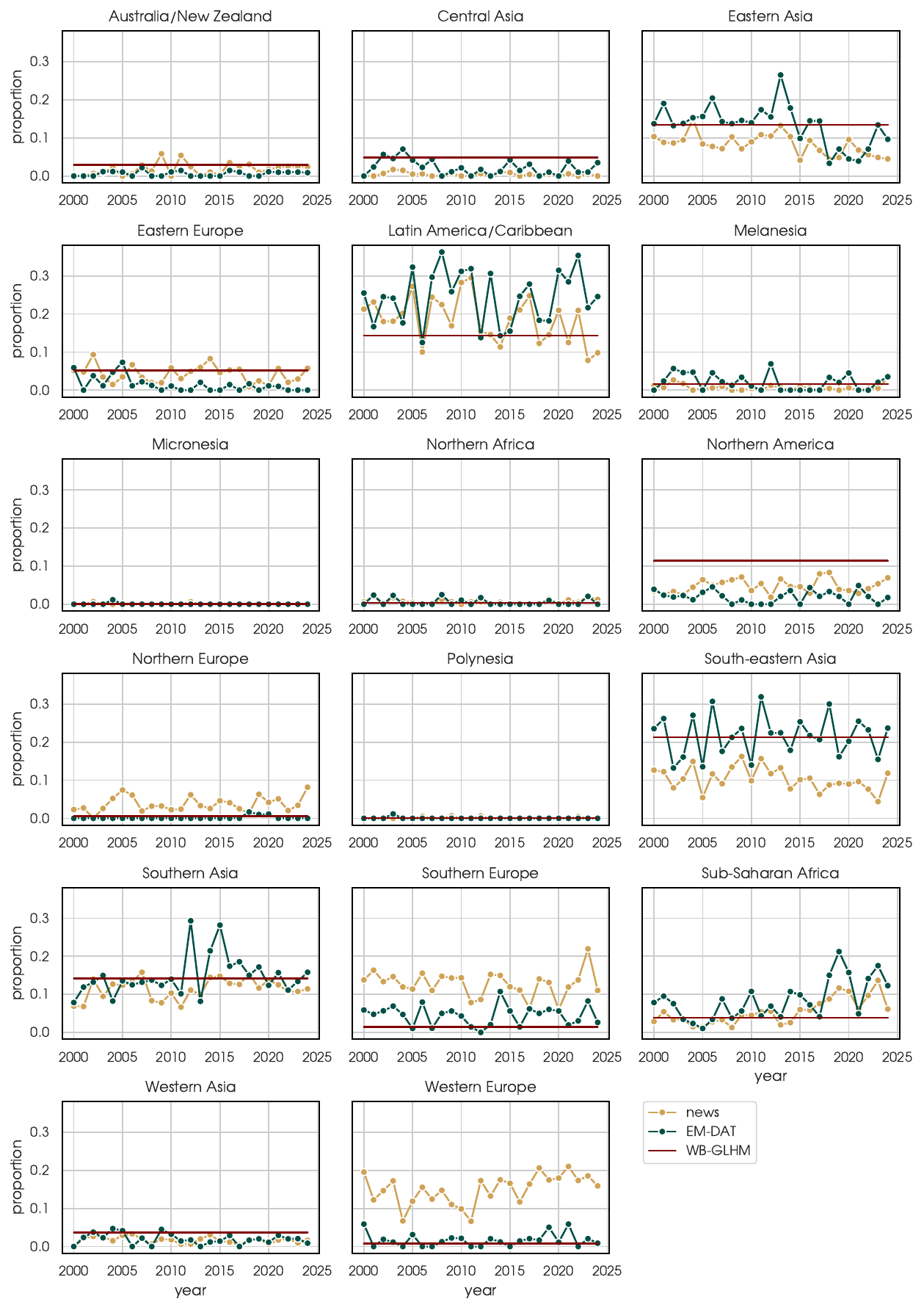}
	\caption{Proportion of (news) events by subregion for each year.}
	\label{fig:years-subregions}
\end{figure*}

Figures \ref{fig:news-events} and \ref{fig:active-days} show the number of identified news events and active days per country (complementing Figure \ref{fig:countmap}). Figure \ref{fig:scatter-docs} shows the almost linear relation between document and news events; generally, more documents result in more events, but a few exceptions occur, e.g.~countries with just one event that encompass a larger number of documents. Figure \ref{fig:years-box-emdat} shows EM-DAT yearly distributions as a counterpart of Figure \ref*{fig:years-box-wb}. The subplots in Figure \ref{fig:years-subregions} represent the evolution of the proportion of documents per subregion over the time period.

Table \ref{tab:valid-subregion} extends the information in Table \ref{tab:valid} with subregions. Finally, Table \ref{tab:full-scores} lists the computed divergence scores for all countries in the sample.

\begin{table*}
	\centering
	\addtolength{\tabcolsep}{-0.2em}
	\begin{tabular}{r|rr|rr}
		\toprule
		&  \multicolumn{2}{c|}{\textbf{news}}  & \textbf{\% EM- }& \textbf{\% WB-}\\
		\textbf{subregion} & \textbf{$n$} & \textbf{\%} & \textbf{DAT} & \textbf{GLHM}\\
		\midrule
		
		Northern Africa & 27 & 0.59 & 0.48 & 0.34 \\
		Sub-Saharan Africa & 260 & 5.69 & 8.90 & 3.83 \\
		Latin America/Caribbean & 844 & 18.48 & 25.25 & 14.31 \\
		Northern America & 226 & 4.95 & 1.89 & 11.42 \\
		Central Asia & 17 & 0.37 & 2.37 & 4.88 \\
		Eastern Asia & 364 & 7.97 & 12.58 & 13.46 \\
		South-eastern Asia & 462 & 10.12 & 21.09 & 21.30 \\
		Southern Asia & 532 & 11.65 & 15.05 & 14.20 \\
		Western Asia & 75 & 1.64 & 2.03 & 3.69 \\
		Eastern Europe & 194 & 4.25 & 1.31 & 5.18 \\
		Northern Europe & 173 & 3.79 & 0.15 & 0.54 \\
		Southern Europe & 580 & 12.70 & 4.45 & 1.46 \\
		Western Europe & 692 & 15.15 & 1.64 & 0.78 \\
		Australia/New Zealand & 83 & 1.82 & 0.68 & 2.93 \\
		Melanesia & 32 & 0.70 & 2.03 & 1.58 \\
		Micronesia & 2 & 0.04 & 0.05 & 0.03 \\
		Polynesia & 4 & 0.09 & 0.05 & 0.06 \\
		
		\cline{1-5}
		\bottomrule
	\end{tabular}
	\caption{Comparison of the number and percentage of identified news events to external measures (subregions).}
	\label{tab:valid-subregion}
\end{table*}

\clearpage
\onecolumn

{
\scriptsize

\begin{longtable}{lrrrrrrr}
	\toprule
	&  &  &  & \textbf{divergence} & \textbf{divergence} & \textbf{binned score} & \textbf{binned score} \\
	\textbf{country} & \textbf{documents} & \textbf{active days} & \textbf{news events} & \textbf{EM-DAT} &\textbf{WB-GLHM} & \textbf{EM-DAT} & \textbf{WB-GLHM} \\
	\midrule
		
		ABW & 1 & 1 & 1 & -0.063 & -0.019 & similar scores & similar scores \\
		AFG & 600 & 61 & 38 & -0.134 & 0.008 & similar scores & similar scores \\
		AGO & 1 & 1 & 1 & -0.186 & -0.229 & lower in news & lower in news \\
		AIA & 0 & 0 & 0 & -0.063 & -0.020 & similar scores & similar scores \\
		ALB & 2 & 2 & 2 & -0.188 & -0.235 & lower in news & lower in news \\
		AND & 0 & 0 & 0 & -0.063 & -0.020 & similar scores & similar scores \\
		ARE & 0 & 0 & 0 & -0.063 & -0.154 & similar scores & lower in news \\
		ARG & 55 & 14 & 10 & 0.020 & -0.115 & similar scores & similar scores \\
		ARM & 0 & 0 & 0 & -0.063 & -0.287 & similar scores & lower in news \\
		ASM & 0 & 0 & 0 & -0.063 & -0.020 & similar scores & similar scores \\
		ATF & 0 & 0 & 0 & -0.063 & -0.020 & similar scores & similar scores \\
		ATG & 0 & 0 & 0 & -0.063 & -0.020 & similar scores & similar scores \\
		AUS & 264 & 56 & 37 & 0.319 & -0.003 & higher in news & similar scores \\
		AUT & 3,617 & 521 & 237 & 0.543 & 0.507 & higher in news & higher in news \\
		AZE & 0 & 0 & 0 & -0.258 & -0.383 & lower in news & lower in news \\
		BDI & 22 & 10 & 7 & -0.114 & 0.141 & similar scores & similar scores \\
		BEL & 20 & 9 & 6 & 0.126 & 0.216 & similar scores & higher in news \\
		BEN & 0 & 0 & 0 & -0.063 & -0.025 & similar scores & similar scores \\
		BES & 0 & 0 & 0 & -0.063 & -0.020 & similar scores & similar scores \\
		BFA & 8 & 2 & 2 & 0.057 & 0.101 & similar scores & similar scores \\
		BGD & 372 & 63 & 42 & 0.024 & 0.054 & similar scores & similar scores \\
		BGR & 7 & 6 & 5 & 0.022 & 0.004 & similar scores & similar scores \\
		BHR & 0 & 0 & 0 & -0.063 & -0.020 & similar scores & similar scores \\
		BHS & 2 & 2 & 1 & -0.063 & -0.025 & similar scores & similar scores \\
		BIH & 206 & 40 & 17 & 0.022 & 0.155 & similar scores & higher in news \\
		BLM & 0 & 0 & 0 & -0.063 & -0.020 & similar scores & similar scores \\
		BLR & 0 & 0 & 0 & -0.063 & -0.020 & similar scores & similar scores \\
		BLZ & 0 & 0 & 0 & -0.063 & -0.166 & similar scores & lower in news \\
		BMU & 0 & 0 & 0 & -0.063 & -0.020 & similar scores & similar scores \\
		BOL & 98 & 35 & 22 & -0.081 & 0.005 & similar scores & similar scores \\
		BRA & 3,056 & 356 & 138 & 0.020 & 0.183 & similar scores & higher in news \\
		BRB & 0 & 0 & 0 & -0.063 & -0.020 & similar scores & similar scores \\
		BRN & 0 & 0 & 0 & -0.063 & -0.096 & similar scores & similar scores \\
		BTN & 42 & 14 & 5 & 0.094 & -0.109 & similar scores & similar scores \\
		BVT & 0 & 0 & 0 & -0.063 & -0.020 & similar scores & similar scores \\
		BWA & 0 & 0 & 0 & -0.063 & -0.020 & similar scores & similar scores \\
		CAF & 5 & 2 & 1 & -0.063 & -0.137 & similar scores & similar scores \\
		CAN & 182 & 43 & 27 & 0.510 & -0.091 & higher in news & similar scores \\
		CCK & 0 & 0 & 0 & -0.063 & -0.020 & similar scores & similar scores \\
		CHE & 3,640 & 638 & 306 & 0.564 & 0.592 & higher in news & higher in news \\
		CHL & 306 & 66 & 45 & 0.131 & 0.079 & similar scores & similar scores \\
		CHN & 4,438 & 580 & 217 & -0.039 & 0.120 & similar scores & similar scores \\
		CIV & 14 & 3 & 2 & -0.332 & -0.045 & lower in news & similar scores \\
		CMR & 42 & 26 & 18 & 0.050 & 0.029 & similar scores & similar scores \\
		COD & 312 & 51 & 36 & -0.017 & 0.204 & similar scores & higher in news \\
		COG & 2 & 1 & 1 & -0.309 & -0.241 & lower in news & lower in news \\
		COK & 0 & 0 & 0 & -0.063 & -0.020 & similar scores & similar scores \\
		COL & 1,161 & 150 & 100 & -0.012 & 0.177 & similar scores & higher in news \\
		COM & 1 & 1 & 1 & -0.186 & -0.307 & lower in news & lower in news \\
		CPV & 0 & 0 & 0 & -0.186 & -0.020 & lower in news & similar scores \\
		CRI & 91 & 22 & 16 & -0.103 & -0.022 & similar scores & similar scores \\
		CUB & 27 & 11 & 6 & -0.096 & -0.010 & similar scores & similar scores \\
		CUW & 0 & 0 & 0 & -0.063 & -0.020 & similar scores & similar scores \\
		CXR & 0 & 0 & 0 & -0.063 & -0.020 & similar scores & similar scores \\
		CYM & 3 & 2 & 2 & 0.057 & 0.101 & similar scores & similar scores \\
		CYP & 0 & 0 & 0 & -0.063 & -0.096 & similar scores & similar scores \\
		CZE & 158 & 96 & 74 & 0.491 & 0.520 & higher in news & higher in news \\
		DJI & 0 & 0 & 0 & -0.186 & -0.198 & lower in news & lower in news \\
		DMA & 57 & 11 & 7 & 0.152 & 0.319 & similar scores & higher in news \\
		DNK & 226 & 22 & 11 & 0.354 & 0.398 & higher in news & higher in news \\
		DOM & 185 & 28 & 17 & -0.062 & 0.209 & similar scores & higher in news \\
		DZA & 40 & 8 & 4 & -0.068 & -0.125 & similar scores & similar scores \\
		ECU & 287 & 66 & 44 & 0.025 & 0.116 & similar scores & similar scores \\
		EGY & 105 & 17 & 9 & 0.319 & 0.069 & higher in news & similar scores \\
		ERI & 1 & 1 & 1 & -0.063 & -0.267 & similar scores & lower in news \\
		ESH & 0 & 0 & 0 & -0.063 & -0.020 & similar scores & similar scores \\
		ESP & 974 & 173 & 116 & 0.518 & 0.380 & higher in news & higher in news \\
		EST & 0 & 0 & 0 & -0.063 & -0.020 & similar scores & similar scores \\
		ETH & 176 & 21 & 13 & -0.042 & -0.058 & similar scores & similar scores \\
		FIN & 0 & 0 & 0 & -0.063 & -0.096 & similar scores & similar scores \\
		FJI & 22 & 4 & 4 & -0.191 & -0.107 & lower in news & similar scores \\
		FLK & 0 & 0 & 0 & -0.063 & -0.020 & similar scores & similar scores \\
		FRA & 953 & 175 & 120 & 0.278 & 0.361 & higher in news & higher in news \\
		FRO & 0 & 0 & 0 & -0.063 & -0.020 & similar scores & similar scores \\
		FSM & 2 & 2 & 1 & -0.063 & -0.019 & similar scores & similar scores \\
		GAB & 0 & 0 & 0 & -0.063 & -0.440 & similar scores & lower in news \\
		GBR & 443 & 103 & 67 & 0.668 & 0.421 & higher in news & higher in news \\
		GEO & 114 & 15 & 10 & -0.103 & -0.054 & similar scores & similar scores \\
		GGY & 0 & 0 & 0 & -0.063 & -0.020 & similar scores & similar scores \\
		GHA & 6 & 6 & 4 & 0.178 & 0.104 & higher in news & similar scores \\
		GIB & 0 & 0 & 0 & -0.063 & -0.020 & similar scores & similar scores \\
		GIN & 11 & 1 & 1 & -0.063 & -0.298 & similar scores & lower in news \\
		GLP & 0 & 0 & 0 & -0.063 & -0.025 & similar scores & similar scores \\
		GMB & 1 & 1 & 1 & -0.063 & -0.019 & similar scores & similar scores \\
		GNB & 0 & 0 & 0 & -0.063 & -0.020 & similar scores & similar scores \\
		GNQ & 0 & 0 & 0 & -0.063 & -0.257 & similar scores & lower in news \\
		GRC & 546 & 87 & 61 & 0.227 & 0.278 & higher in news & higher in news \\
		GRD & 0 & 0 & 0 & -0.186 & -0.020 & lower in news & similar scores \\
		GRL & 31 & 13 & 8 & 0.298 & 0.091 & higher in news & similar scores \\
		GTM & 1,008 & 122 & 55 & -0.011 & 0.241 & similar scores & higher in news \\
		GUF & 0 & 0 & 0 & -0.063 & -0.096 & similar scores & similar scores \\
		GUM & 0 & 0 & 0 & -0.063 & -0.025 & similar scores & similar scores \\
		GUY & 0 & 0 & 0 & -0.063 & -0.206 & similar scores & lower in news \\
		HKG & 39 & 3 & 3 & 0.128 & 0.124 & similar scores & similar scores \\
		HMD & 0 & 0 & 0 & -0.063 & -0.020 & similar scores & similar scores \\
		HND & 427 & 53 & 26 & 0.036 & 0.156 & similar scores & higher in news \\
		HRV & 24 & 14 & 9 & 0.319 & 0.078 & higher in news & similar scores \\
		HTI & 503 & 92 & 48 & 0.040 & 0.379 & similar scores & higher in news \\
		HUN & 72 & 18 & 9 & 0.319 & 0.264 & higher in news & higher in news \\
		IDN & 3,059 & 291 & 136 & -0.110 & 0.128 & similar scores & similar scores \\
		IMN & 0 & 0 & 0 & -0.063 & -0.020 & similar scores & similar scores \\
		IND & 2,853 & 382 & 173 & 0.034 & 0.127 & similar scores & similar scores \\
		IOT & 0 & 0 & 0 & -0.063 & -0.020 & similar scores & similar scores \\
		IRL & 11 & 3 & 2 & 0.057 & -0.016 & similar scores & similar scores \\
		IRN & 38 & 15 & 10 & -0.052 & -0.263 & similar scores & lower in news \\
		IRQ & 4 & 2 & 2 & 0.057 & -0.248 & similar scores & lower in news \\
		ISL & 7 & 5 & 4 & 0.178 & -0.069 & higher in news & similar scores \\
		ISR & 12 & 6 & 6 & 0.248 & 0.193 & higher in news & higher in news \\
		ITA & 4,991 & 631 & 314 & 0.283 & 0.488 & higher in news & higher in news \\
		JAM & 45 & 11 & 10 & 0.020 & 0.165 & similar scores & higher in news \\
		JEY & 0 & 0 & 0 & -0.063 & -0.020 & similar scores & similar scores \\
		JOR & 1 & 1 & 1 & -0.063 & -0.166 & similar scores & lower in news \\
		JPN & 3,045 & 275 & 114 & 0.029 & 0.188 & similar scores & higher in news \\
		KAZ & 21 & 4 & 3 & 0.128 & -0.359 & similar scores & lower in news \\
		KEN & 73 & 32 & 18 & -0.073 & 0.165 & similar scores & higher in news \\
		KGZ & 119 & 99 & 8 & -0.181 & -0.308 & lower in news & lower in news \\
		KHM & 7 & 4 & 4 & 0.178 & -0.311 & higher in news & lower in news \\
		KIR & 1 & 1 & 1 & -0.063 & -0.019 & similar scores & similar scores \\
		KNA & 0 & 0 & 0 & -0.063 & -0.020 & similar scores & similar scores \\
		KOR & 227 & 33 & 16 & -0.072 & 0.014 & similar scores & similar scores \\
		KWT & 0 & 0 & 0 & -0.063 & -0.020 & similar scores & similar scores \\
		LAO & 23 & 4 & 4 & -0.068 & -0.318 & similar scores & lower in news \\
		LBN & 3 & 1 & 1 & -0.063 & -0.207 & similar scores & lower in news \\
		LBR & 5 & 1 & 1 & -0.063 & -0.462 & similar scores & lower in news \\
		LBY & 26 & 2 & 1 & -0.063 & -0.188 & similar scores & lower in news \\
		LCA & 0 & 0 & 0 & -0.186 & -0.020 & lower in news & similar scores \\
		LIE & 2 & 2 & 2 & 0.057 & 0.101 & similar scores & similar scores \\
		LKA & 581 & 66 & 33 & -0.075 & 0.490 & similar scores & higher in news \\
		LSO & 2 & 2 & 2 & 0.057 & 0.054 & similar scores & similar scores \\
		LTU & 0 & 0 & 0 & -0.063 & -0.020 & similar scores & similar scores \\
		LUX & 39 & 16 & 15 & 0.408 & 0.446 & higher in news & higher in news \\
		LVA & 1 & 1 & 1 & -0.063 & -0.019 & similar scores & similar scores \\
		MAC & 0 & 0 & 0 & -0.063 & -0.020 & similar scores & similar scores \\
		MAF & 0 & 0 & 0 & -0.063 & -0.020 & similar scores & similar scores \\
		MAR & 73 & 14 & 9 & 0.073 & 0.038 & similar scores & similar scores \\
		MCO & 0 & 0 & 0 & -0.063 & -0.020 & similar scores & similar scores \\
		MDA & 2 & 2 & 2 & 0.057 & 0.101 & similar scores & similar scores \\
		MDG & 121 & 27 & 16 & -0.006 & -0.105 & similar scores & similar scores \\
		MDV & 0 & 0 & 0 & -0.063 & -0.020 & similar scores & similar scores \\
		MEX & 1,170 & 138 & 61 & -0.052 & 0.007 & similar scores & similar scores \\
		MHL & 0 & 0 & 0 & -0.063 & -0.020 & similar scores & similar scores \\
		MKD & 25 & 1 & 1 & -0.348 & -0.154 & lower in news & lower in news \\
		MLI & 0 & 0 & 0 & -0.063 & -0.067 & similar scores & similar scores \\
		MLT & 1 & 1 & 1 & -0.063 & -0.019 & similar scores & similar scores \\
		MMR & 636 & 51 & 28 & -0.031 & -0.131 & similar scores & similar scores \\
		MNE & 34 & 5 & 4 & 0.178 & 0.012 & higher in news & similar scores \\
		MNG & 1 & 1 & 1 & -0.186 & -0.500 & lower in news & lower in news \\
		MNP & 0 & 0 & 0 & -0.063 & -0.020 & similar scores & similar scores \\
		MOZ & 34 & 7 & 6 & 0.054 & 0.032 & similar scores & similar scores \\
		MRT & 1 & 1 & 1 & -0.063 & -0.067 & similar scores & similar scores \\
		MSR & 0 & 0 & 0 & -0.063 & -0.020 & similar scores & similar scores \\
		MTQ & 1 & 1 & 1 & -0.186 & -0.019 & lower in news & similar scores \\
		MUS & 1 & 1 & 1 & -0.063 & -0.137 & similar scores & similar scores \\
		MWI & 90 & 14 & 5 & 0.022 & 0.082 & similar scores & similar scores \\
		MYS & 58 & 23 & 17 & -0.025 & -0.160 & similar scores & lower in news \\
		MYT & 0 & 0 & 0 & -0.063 & -0.020 & similar scores & similar scores \\
		NAM & 0 & 0 & 0 & -0.063 & -0.229 & similar scores & lower in news \\
		NCL & 0 & 0 & 0 & -0.063 & -0.137 & similar scores & similar scores \\
		NER & 0 & 0 & 0 & -0.063 & -0.119 & similar scores & similar scores \\
		NFK & 0 & 0 & 0 & -0.063 & -0.020 & similar scores & similar scores \\
		NGA & 2 & 2 & 2 & -0.137 & -0.192 & similar scores & lower in news \\
		NIC & 314 & 64 & 38 & 0.180 & 0.280 & higher in news & higher in news \\
		NLD & 10 & 6 & 6 & 0.248 & 0.292 & higher in news & higher in news \\
		NOR & 962 & 118 & 80 & 0.576 & 0.294 & higher in news & higher in news \\
		NPL & 1,662 & 312 & 142 & 0.102 & 0.246 & similar scores & higher in news \\
		NRU & 0 & 0 & 0 & -0.063 & -0.020 & similar scores & similar scores \\
		NZL & 344 & 66 & 46 & 0.195 & 0.064 & higher in news & similar scores \\
		OMN & 0 & 0 & 0 & -0.186 & -0.344 & lower in news & lower in news \\
		PAK & 890 & 172 & 89 & 0.032 & 0.227 & similar scores & higher in news \\
		PAN & 83 & 22 & 16 & -0.103 & 0.072 & similar scores & similar scores \\
		PCN & 0 & 0 & 0 & -0.063 & -0.020 & similar scores & similar scores \\
		PER & 733 & 166 & 103 & 0.068 & 0.185 & similar scores & higher in news \\
		PHL & 4,244 & 411 & 153 & -0.062 & 0.144 & similar scores & similar scores \\
		PLW & 0 & 0 & 0 & -0.063 & -0.096 & similar scores & similar scores \\
		PNG & 689 & 37 & 20 & -0.098 & -0.069 & similar scores & similar scores \\
		POL & 118 & 35 & 23 & 0.482 & 0.324 & higher in news & higher in news \\
		PRI & 2 & 2 & 2 & -0.228 & -0.045 & lower in news & similar scores \\
		PRK & 36 & 18 & 13 & 0.038 & 0.102 & similar scores & similar scores \\
		PRT & 316 & 39 & 30 & 0.184 & 0.297 & higher in news & higher in news \\
		PRY & 5 & 3 & 3 & 0.005 & 0.166 & similar scores & higher in news \\
		PSE & 0 & 0 & 0 & -0.063 & -0.096 & similar scores & similar scores \\
		PYF & 0 & 0 & 0 & -0.063 & -0.247 & similar scores & lower in news \\
		QAT & 0 & 0 & 0 & -0.063 & -0.020 & similar scores & similar scores \\
		REU & 0 & 0 & 0 & -0.186 & -0.301 & lower in news & lower in news \\
		ROU & 68 & 27 & 20 & 0.069 & 0.112 & similar scores & similar scores \\
		RUS & 196 & 66 & 53 & 0.202 & -0.044 & higher in news & similar scores \\
		RWA & 36 & 14 & 11 & -0.168 & 0.188 & lower in news & higher in news \\
		SAU & 3 & 3 & 2 & 0.057 & -0.189 & similar scores & lower in news \\
		SDN & 0 & 0 & 0 & -0.063 & -0.229 & similar scores & lower in news \\
		SEN & 0 & 0 & 0 & -0.063 & -0.020 & similar scores & similar scores \\
		SGP & 0 & 0 & 0 & -0.063 & -0.020 & similar scores & similar scores \\
		SGS & 0 & 0 & 0 & -0.063 & -0.025 & similar scores & similar scores \\
		SHN & 0 & 0 & 0 & -0.063 & -0.020 & similar scores & similar scores \\
		SJM & 0 & 0 & 0 & -0.063 & -0.025 & similar scores & similar scores \\
		SLB & 16 & 3 & 3 & -0.157 & -0.245 & similar scores & lower in news \\
		SLE & 384 & 34 & 18 & 0.122 & 0.241 & similar scores & higher in news \\
		SLV & 512 & 75 & 32 & 0.027 & 0.336 & similar scores & higher in news \\
		SMR & 0 & 0 & 0 & -0.063 & -0.020 & similar scores & similar scores \\
		SOM & 13 & 4 & 3 & 0.128 & -0.084 & similar scores & similar scores \\
		SPM & 0 & 0 & 0 & -0.063 & -0.020 & similar scores & similar scores \\
		SRB & 112 & 20 & 11 & 0.036 & 0.137 & similar scores & similar scores \\
		SSD & 0 & 0 & 0 & -0.063 & -0.166 & similar scores & lower in news \\
		STP & 0 & 0 & 0 & -0.063 & -0.020 & similar scores & similar scores \\
		SUR & 3 & 1 & 1 & -0.063 & -0.137 & similar scores & similar scores \\
		SVK & 89 & 7 & 4 & 0.178 & -0.063 & higher in news & similar scores \\
		SVN & 392 & 31 & 13 & 0.137 & 0.159 & similar scores & higher in news \\
		SWE & 162 & 12 & 8 & 0.298 & 0.100 & higher in news & similar scores \\
		SWZ & 0 & 0 & 0 & -0.063 & -0.067 & similar scores & similar scores \\
		SXM & 0 & 0 & 0 & -0.063 & -0.020 & similar scores & similar scores \\
		SYC & 13 & 6 & 2 & 0.057 & 0.101 & similar scores & similar scores \\
		SYR & 6 & 3 & 2 & 0.057 & -0.102 & similar scores & similar scores \\
		TCA & 0 & 0 & 0 & -0.063 & -0.020 & similar scores & similar scores \\
		TCD & 0 & 0 & 0 & -0.063 & -0.229 & similar scores & lower in news \\
		TGO & 0 & 0 & 0 & -0.063 & -0.025 & similar scores & similar scores \\
		THA & 177 & 31 & 18 & -0.062 & -0.039 & similar scores & similar scores \\
		TJK & 12 & 6 & 6 & -0.360 & -0.301 & lower in news & lower in news \\
		TKL & 0 & 0 & 0 & -0.063 & -0.020 & similar scores & similar scores \\
		TKM & 0 & 0 & 0 & -0.063 & -0.298 & similar scores & lower in news \\
		TLS & 61 & 5 & 1 & -0.258 & -0.206 & lower in news & lower in news \\
		TON & 6 & 5 & 1 & -0.063 & -0.025 & similar scores & similar scores \\
		TTO & 1 & 1 & 1 & -0.309 & -0.067 & lower in news & similar scores \\
		TUN & 4 & 4 & 4 & 0.178 & 0.104 & higher in news & similar scores \\
		TUR & 300 & 63 & 43 & 0.123 & -0.021 & similar scores & similar scores \\
		TUV & 0 & 0 & 0 & -0.063 & -0.020 & similar scores & similar scores \\
		TWN & 530 & 84 & 47 & 0.094 & 0.023 & similar scores & similar scores \\
		TZA & 84 & 25 & 15 & 0.063 & 0.108 & similar scores & similar scores \\
		UGA & 236 & 47 & 36 & 0.012 & 0.402 & similar scores & higher in news \\
		UKR & 4 & 4 & 4 & 0.178 & -0.082 & higher in news & similar scores \\
		URY & 11 & 3 & 3 & 0.128 & 0.166 & similar scores & higher in news \\
		USA & 2,172 & 356 & 191 & 0.206 & 0.116 & higher in news & similar scores \\
		UZB & 0 & 0 & 0 & -0.186 & -0.426 & lower in news & lower in news \\
		VAT & 1 & 1 & 1 & -0.063 & -0.019 & similar scores & similar scores \\
		VCT & 7 & 4 & 2 & -0.137 & 0.101 & similar scores & similar scores \\
		VEN & 294 & 54 & 33 & 0.137 & 0.097 & similar scores & similar scores \\
		VGB & 0 & 0 & 0 & -0.063 & -0.020 & similar scores & similar scores \\
		VIR & 0 & 0 & 0 & -0.063 & -0.020 & similar scores & similar scores \\
		VNM & 673 & 96 & 54 & -0.086 & 0.035 & similar scores & similar scores \\
		VUT & 43 & 8 & 5 & -0.101 & -0.102 & similar scores & similar scores \\
		WLF & 0 & 0 & 0 & -0.063 & -0.020 & similar scores & similar scores \\
		WSM & 6 & 4 & 3 & 0.128 & 0.172 & similar scores & higher in news \\
		YEM & 67 & 11 & 8 & -0.046 & -0.064 & similar scores & similar scores \\
		ZAF & 406 & 46 & 28 & 0.271 & 0.192 & higher in news & higher in news \\
		ZMB & 3 & 2 & 2 & 0.057 & 0.002 & similar scores & similar scores \\
		ZWE & 32 & 6 & 4 & 0.055 & 0.104 & similar scores & similar scores \\
	\bottomrule
	\caption{Computed measures for all countries.}
	\label{tab:full-scores}
\end{longtable}

}

\end{document}